\documentclass[journal,10pt]{IEEEtran}

\usepackage{ifpdf}
\usepackage{amsmath}
\usepackage{array}
\usepackage{fixltx2e}
\usepackage{stfloats}
\usepackage{url}

\usepackage{graphicx,psfrag,epsf}
\usepackage{mathtools}
\usepackage{booktabs}
\usepackage[english]{babel} 
\usepackage{amsmath,amsfonts,amsthm}
\usepackage{chngcntr}
\usepackage{ amssymb }
\usepackage{array}
\usepackage{booktabs} 
\usepackage{setspace}
\usepackage{fix-cm}
\usepackage{xcolor}
\usepackage[normalem]{ulem}
\usepackage{hyperref}
\hypersetup{colorlinks,urlcolor=blue}
\usepackage{cite}

\usepackage{url}
\usepackage{tikz}
\usepackage{cool}
\usepackage{enumitem}
\usepackage{soul}
\usepackage{threeparttable}
\usepackage{tablefootnote}
\usepackage{color}
\usepackage{tabularx}
\usepackage{caption} 
\usepackage{footnote}
\usepackage{algorithm}
\usepackage[noend]{algpseudocode}
\usepackage{footnote}
\usepackage{multirow}

\makeatletter
\DeclareUrlCommand\ULurl@@{%
  \def\UrlLeft{\bgroup}%
  \def\UrlRight{\egroup}}
\def\ULurl@#1{\hyper@linkurl{\ULurl@@{#1}}{#1}}
\DeclareRobustCommand*\ULurl{\hyper@normalise\ULurl@}

\newcommand{\distas}[1]{\mathbin{\overset{#1}{\kern\z@\sim}}}%

\newsavebox{\mybox}\newsavebox{\mysim}
\newcommand{\distras}[1]{%
  \savebox{\mybox}{\hbox{\kern3pt$\scriptstyle#1$\kern3pt}}%
  \savebox{\mysim}{\hbox{$\sim$}}%
  \mathbin{\overset{#1}{\kern\z@\resizebox{\wd\mybox}{\ht\mysim}{$\sim$}}}%
}
\DeclareMathOperator*{\argmin}{\arg\!\min}
\def\bbr{{\Bbb{R}}} 
\def\bbe{{\Bbb{E}}} 

\def\bbs{{\Bbb{S}}}
\def\bbd{{\Bbb{D}}}

\def\bbv{{\Bbb{V}}}
\def\bbw{{\Bbb{W}}}

\newcommand{\half}{ \mbox{\small$\frac{1}{2}$}}

\newcommand{\be}{\begin{equation}}
\newcommand{\ee}{\end{equation}}

\def\w{\omega}
\def\e{\varepsilon}
\def\O{\Omega}

\def\vect {{\rm vec}}

\def\tr {{\rm tr}}
\def\rank{{\rm rank}}
\def\dim {{\rm dim}}

\def\sol {{\rm Sol}}

\newcommand{\V}{{\cal V}}

\newcommand{\F}{{\cal F}}

\newcommand{\A}{{\cal A}}

\newcommand{\G}{{\cal G}}

\newcommand{\M}{{\cal M}}

\newcommand{\LL}{{\cal L}}

\newcommand{\W}{\mathcal{W}}
\newcommand{\s}{\mathcal{S}}

\newcommand{\cS}{{\mathfrak S}}

\newcommand{\cF}{{\mathfrak F}}

\newcommand{\cR}{{\mathfrak R}}

\newcommand{\cf}{{\mathfrak f}}

\newcommand{\N}{{\cal N}}

\newcommand{\T}{{\cal T}}

\newtheorem{theorem}{Theorem}[section]

\newtheorem{proposition}{Proposition}[section]
\newtheorem{example}{Example}[section]
\newtheorem{remark}{Remark}[section]
\newtheorem{definition}{Definition}[section]

\newtheorem{assu}{Assumption}[section]

\def\yao{\textcolor{black}}
\def\alex{\textcolor{black}}

\begin{document}
\title{Matrix completion  with deterministic  pattern: \\A geometric perspective}
\author{Alexander Shapiro, ~
        Yao~Xie, ~Rui Zhang
\thanks{Alexander Shapiro (e-mail: \ULurl{ashapiro@isye.gatech.edu}), Yao Xie (e-mail: \ULurl{yao.xie@isye.gatech.edu}) and Rui Zhang (e-mail: \ULurl{ruizhang_ray@gatech.edu}) are with the H. Milton Stewart School of Industrial and Systems Engineering, Georgia Institute of Technology, Atlanta, GA.}
\thanks{Research of   Alexander Shapiro was partly supported by NSF grant 1633196 and  DARPA EQUiPS program, grant SNL 014150709. Research of Yao Xie was partially supported by NSF grants CCF-1442635, CMMI-1538746, an NSF CAREER
Award CCF-1650913, and a S.F. Express award.}
}



\maketitle

\begin{abstract}
We consider the matrix completion problem with a  deterministic pattern of observed entries. In this setting, we aim to answer the question: under what condition there will be (at least locally) unique solution to the matrix completion problem, i.e., the underlying true matrix is identifiable. We answer the question from a certain  point of view and  outline  a geometric perspective. We give an algebraically verifiable sufficient condition, which we call the {\it well-posedness condition},  for the local uniqueness of MRMC solutions. \yao{We argue that this condition is necessary for local stability of MRMC solutions, and we show that the condition is generic using the characteristic rank.} We also argue that the low-rank approximation approaches are more stable than MRMC and further propose a sequential statistical testing procedure to determine the ``true" rank   from observed entries. Finally, we provide  numerical examples aimed at verifying  validity of the presented  theory.
\end{abstract}

\IEEEpeerreviewmaketitle

\section{Introduction}

Matrix completion (e.g., \cite{candes2009exact,recht2010guaranteed,candes2010power}) is a fundamental problem in signal processing and machine learning, which studies the recovery of a low-rank matrix from an observation of a subset of its entries. It has attracted a lot attention from researchers and practitioners and there are  various motivating real-world applications including recommender systems and the Netflix challenge (see a recent overview in \cite{DavenportRomberg16}).
A popular approach for matrix completion is to find a matrix of minimal rank satisfying the observation constraints. Due to the non-convexity of the rank function, popular approaches are convex relaxation (see, e.g., \cite{Fazel}) and nuclear norm minimization. There is a rich literature, both in establishing performance bounds, developing efficient algorithms and providing performance guarantees. Recently there has also been new various results for non-convex formulations of matrix completion problem  (see, e.g.,  \cite{Hsieh2017}).

Existing conditions ensuring recovery of the minimal rank matrix are usually formulated in terms of missing-at-random entries and under an assumption of the so-called bounded-coherence (see a survey for other approaches in \cite{DavenportRomberg16}; we do not aim to give a complete overview of the vast literature). These results are typically aimed at establishing the recovery with a  high probability. \yao{In addition, there has been much work on low-rank matrix recovery (see, e.g.,} \cite{Eldar2012}, which studies a related problem: the uniqueness conditions for minimum rank matrix recovery with {\it random} linear measurements of the true matrix; here the linear measurements correspond  to inner product of a measurement mask matrix with the true matrix, and hence, the observations are different from that in matrix completion).

With a deterministic  pattern of observed entries, a complete characterization of the identifiable matrix for matrix completion remains an important yet open question: under what conditions for the pattern, there will be (at least locally) unique solution?  Recent work \cite{Nowak16} provides insights into this problem by studying the so-called completable problems and establishing conditions ensuring the existence of at most finitely many rank-$r$ matrices that agree with all its observed entries.  \yao{A related work \cite{AggarwalWang2017} studied this problem when there is a sparse noise that corrupts the entries. The rank estimation problem has been discussed in \cite{Pimentel2016, Ashraphijuo2017}, and related tensor completion problem in \cite{Ashraphijuo2017_2}: the goal in these works are different though; they aim to find upper and lower bound for the true rank, whereas our rank selection test in Section \ref{sec-stat} determines the most plausible rank from a statistical point of view.}

In this paper, we aim to answer the question from a somewhat different point of view and to give a geometric perspective. In particular, we consider the solution of the  Minimum Rank Matrix Completion (MRMC) formulation, which leads to a non-convex optimization problem. We address the following questions: (i) Given observed entries arranged according to a   (deterministic)  pattern, by solving the  MRMC  problem, what is the minimum achievable rank? (ii) Under what conditions, there will be a unique matrix that is a solution to the  MRMC  problem? We give a sufficient   condition  (which we call the {\it well-posedness condition}) for the local uniqueness of MRMC solutions, and illustrate how  such condition can be verified. \yao{We also show that such well-posedness condition is generic using the concept of characteristic rank.} In addition, we also consider the convex relaxation and nuclear norm minimization formulations.

\yao{Based on our theoretical results,} we argue that  given $m$ observations  of an $n_1\times n_2$ matrix, if the minimal rank $r^*$ is less than $\cR(n_1,n_2,m) := (n_1+n_2)/2 - [(n_1+n_2)^2/4 - m]^{1/2}$, then the corresponding solution is unstable in the sense that an arbitrary small perturbation of the observed values  can make this rank unattainable.
On the other hand if  $r^*>  \cR(n_1,n_2,m)$, then almost surely the solution is not (even locally) unique (cf., \cite{AShapiro_2017}).
This indicates that except on rare occasions, the MRMC problem cannot have both properties of possessing unique and stable solutions.
Consequently,  what makes sense is to try to solve the minimum rank problem approximately and hence to consider low-rank approximation approaches (such as an approach mentioned in \cite{DavenportRomberg16,Tropp17}) as a better alternative to the MRMC  formulation.

We also propose a sequential statistical testing procedure to determine the `true' rank   from noisy observed entries. Such statistical approach can be useful for many existing low-rank matrix completion algorithms,  which  require a pre-specification of the matrix rank, such as the alternating minimization approach to solving the non-convex problem by representing the low-rank matrix as a product of two low-rank matrix factors (see, e.g.,  \cite{SL2014,DavenportRomberg16,MaWang2017}).

The paper is organized as follows.  In the next section, we introduce the considered setting and some basic definitions. In Section \ref{sec:mcomp} we present the problem set-up, including the MRMC, LRMA, and convex relaxation formulations. Section \ref{sec:main_results} contains the main theoretical results. 
A statistical test of rank is presented in Section \ref{sec-stat}. In Section \ref{sec-numer} we present numerical results related to the developed theory. Finally Section \ref{sec:conclusion} concludes the paper. All proofs are transferred to the Appendix.

We use   conventional notations.
For $a\in \bbr$ we denote by $\lceil a\rceil$ the least integer  that is greater than or equal to $a$.
 By $A\otimes B$ we denote the Kronecker product of matrices (vectors) $A$ and $B$, and by $\vect(A)$ column vector obtained by stacking columns of matrix $A$.  We use the following matrix identity
for matrices $A,B,C$  of appropriate order
\begin{equation}\label{matreq}
 \vect(ABC)=(C^\top\otimes A)\vect(B).
\end{equation}
By $\bbs^p$ we denote the linear space of $p\times p$ symmetric matrices and by writing $X\succeq 0$ we mean that matrix $X\in \bbs^p$ is positive semidefinite.  By $\sigma_i(Y)$ we denote  the $i$-th largest singular value of matrix $Y\in \bbr^{n_1\times n_2}$. By $I_p$ we denote the  identity matrix of dimension $p$.


\section{Matrix completion and {problem set-up}}
\label{sec:mcomp}



Consider the problem of recovering an $n_1\times n_2$  data matrix   of low rank  when observing  a small number $m$  of its entries, which are denoted as $M_{ij}$, $(i, j) \in \O$. \alex{We assume that $n_1\ge 2$ and $n_2\ge 2$.} Here $ \O\subset \{1,...,n_1\}\times \{1,...,n_2\}$ is an index set  of cardinality $m$.
\yao{The low-rank matrix completion problem, or matrix completion problem, aims to infer the missing entries, based on the available observations $M_{ij}$, $(i, j) \in \O$, by using a matrix whose rank is as small as possible.
}

\yao{Low-rank matrix completion problem is usually studied under a missing-at-random  model, under which the necessary and sufficient conditions for perfect recovery of the true matrix are known \cite{candrecht,CandesTao10,Recht2011,Gross11,Chen2014,ChenBhojanapalli2014}. Study of deterministic sampling pattern is relatively rare. This includes the finitely rank-$r$ completability problem in \cite{Nowak16}, which shows the conditions for the deterministic sampling pattern such that there exists at most {\it finitely many} rank-$r$ matrices that agrees with its observed entries.} 
\yao{In this paper, we study a related but different problem, i.e., when will the matrix have a unique way to be completed, given a fixed sampling pattern. This is a fundamental problem related to {\it the identifiability} of a low-rank matrix  given an observation pattern $\Omega$.}

\subsection{Definitions}

Lt us  introduce some necessary definitions. Denote by $M$ the $n_1\times n_2$ matrix with the specified entries  $M_{ij}$, $(i,j)\in \O$, and all other  entries equal zero.
Consider $\O^c:=\{1,...,n_1\}\times \{1,...,n_2\}\setminus \O$, the complement of the index  set $\O$, and
define
 \[
 \bbv_\O:=\left \{Y\in \bbr^{n_1\times n_2}: Y_{ij}=0,\;(i,j)\in  \O^c\right\}.
 \]
 This linear space  represents the set of matrices that are filled with zeros at the locations of the  unobserved entries.
 Similarly  define
 \[
 \bbv_{\O^c}:=\left \{Y\in \bbr^{n_1\times n_2}: Y_{ij}=0,\;(i,j)\in  \O\right\}.
 \]
%
%
By $P_\O$ we denote the projection onto the space $\bbv_\O$, i.e., $[P_\O(Y)]_{ij}=Y_{ij}$ for $(i,j)\in \O$ and $[P_\O(Y)]_{ij}=0$ for $(i,j)\in \O^c$.
 By this construction, \yao{$\{M + X: X \in \bbv_{\O^c}\}$} is the affine space of all matrices that satisfy the observation constraints.   Note that $M\in \bbv_\O$ and
 the dimension of the linear space $\bbv_\O$ is
$\dim(\bbv_\O)=m$, while
 $\dim(\bbv_{\O^c})=n_1n_2-m$.

We say that a property holds for {\em almost every} (a.e.) $M_{ij}$, or almost surely, if the set of matrices $Y\in \bbv_\O$ for which this property does not hold has Lebesgue measure zero in the space  $\bbv_\O$.
\subsection{Minimum Rank Matrix Completion (MRMC)}

\yao{Since the true rank is unknown, a natural approach is to find the minimum rank matrix that is consistent with the observations.} This goal can be written as the following optimization problem referred to  as the Minimum Rank Matrix Completion (MRMC),
\begin{equation}\label{mcomp-1}
\min_{Y\in \bbr^{n_1\times n_2}} \rank (Y)\;~
 {\rm subject\;to}\;  Y_{ij}=M_{ij}, \;(i,j)\in \O.
\end{equation}

\yao{In general, the rank minimization problem is non-convex and NP-hard to solve.
However, this problem is fundamental to various efficient heuristics derived from here. Largely, there are two categories of approximation heuristics: (i) approximate the rank function with some surrogate function such as the nuclear norm function, (ii) or solve a sequence of rank-constrained problems such as the matrix factorization based method, which we will discuss below. Approach (ii) requires to specify the target rank of the recovered matrix beforehand, which we will present a novel statistical test next. }

\subsection{Low Rank Matrix Approximation (LRMA)}

Consider the problem
\begin{equation}\label{mcomp-2}
 \min_{Y\in \bbr^{n_1\times n_2},\,X\in \bbv_{\O^c}} F(M+X,Y)\;\;{\rm s.t.}\; \rank(Y)=r,
\end{equation}
where $M\in \bbv_{\O}$ is the given data matrix, and
$F(A,B)$ is a discrepancy between matrices $A,B\in\bbr^{n_1\times n_2}$.
For example,
let $F(A,B):=\|A-B\|^2_F$ with  $\|Y\|_F^2=\tr(Y^\top Y)=\sum_{i,j}Y_{ij}^2,$ being the  Frobenius norm. Define   the set of $n_1\times n_2$ matrices of rank $r$
\begin{equation}\label{manifrank}
\M_r:=\left\{Y\in \bbr^{n_1\times n_2}:\rank (Y)=r\right\}
\end{equation} Then \eqref{mcomp-2} becomes the least squares problem
\begin{equation}\label{leastsq}
\min_{Y\in \M_r}\sum_{(i,j)\in \O}\left(M_{ij}-Y_{ij}\right)^2.
\end{equation}
The least squares approach although is natural, is not the only one possible. For example, in the statistical approach to  Factor Analysis the discrepancy function is based on the Maximum Likelihood method and is more involved
{\color{black}(e.g., \cite{browne}).}

 \subsection{SDP formulations: Trace and nuclear norm minimization}

An alternative approach to the  MRMC problem, which has  been studied extensively in the literature,  is the convex relaxation formulation (e.g., \cite{candes2009exact,Fazel}). Let $\cS\subset \{1,...,p\}\times \{1,...,p\}$ be the symmetric  index set corresponding to the index set $\O$, i.e.,  $(i, n_1+j)\in \cS$ when $1\le i\le n_1$,  if and only if $(i,j)\in \Omega$; and if $(i,j)\in \cS$, then $(j,i)\in \cS$. By $\cS^c\subset \{1,...,p\}\times \{1,...,p\}$ we denote the symmetric index set complement of $\cS$. Define
\[
 \bbw_\cS:=\{X\in \bbs^p:X_{ij}=0,\;(i,j)\in \cS^c\}\]
 and\[
 \bbw_{\cS^c}:=\{X\in \bbs^p:X_{ij}=0,\;(i,j)\in \cS\}.
 \]
 Define $\Xi\in \bbs^p$,    $p=n_1+n_2$, a symmetric matrix of the following  form,  \yao{that contains the data},
\[
\Xi=\begin{bmatrix}
           0 & M \\
           M^\top & 0
          \end{bmatrix}.\]
The MRMC problem \eqref{mcomp-1}  can be formulated in the following equivalent form
\begin{equation}\label{sim-1}
 \min_{X\in \bbw_{\cS^c}}  \rank (\Xi+X)\; {\rm subject\;to}\;\Xi+X\succeq 0.
\end{equation}
 Minimization in \eqref{sim-1} is performed over matrices $X\in \bbs^p$ which are complement to $\Xi$ in the sense of  having zero entries  at all places corresponding to  the specified values $M_{ij}$, $(i,j)\in  \O$.
We consider a   more general minimum rank problem of the form \eqref{sim-1} in that we allow the index set $\cS$  to be a general symmetric subset of $ \{1,...,p\}\times \{1,...,p\}$,
with a given  matrix $\Xi\in \bbw_{\cS}$.
Note that $\bbw_\cS\cap\bbw_{\cS^c}=\{0\}$ and  $\bbw_\cS+\bbw_{\cS^c}=\bbs^p$.

As a heuristic it was suggested in \cite{Fazel}   to approximate problem \eqref{sim-1} by the following \yao{trace minimization} problem
\begin{equation}\label{sim-2}
 \min_{X\in \bbw_{\cS^c}}  \tr ( X) \; {\rm subject\;to}\;\Xi+X\succeq 0,
\end{equation}
which is equivalent to the following {\it nuclear norm minimization} problem
\begin{equation}\label{sim-3}
 \min_{X}  \|X+M\|_* \; {\rm subject\;to}\;X\in \bbv_{\O^c}.
\end{equation}
Problem \eqref{sim-2} is a special case of the following general SDP problem (if we introduce a weight matrix $C\in \bbw_{\cS^c}$):
\begin{equation}\label{sim-3}
 \min_{X\in \bbw_{\cS^c}}  \tr(C  X)
  \; {\rm subject\;to}\;\Xi+X\succeq 0.
\end{equation}
\yao{The above formulation is a semidefinite programming (SDP)  problem and can be solved efficiently, e.g., by using the singular value thresholding algorithm \cite{CaiCandesShen2010}. Therefore, it has been commonly adopted as an approximation to the minimum rank problem.}

\section{Main theoretical results}\label{sec:main_results}

 To gain insights into the identifiability issue of matrix completion, we aim to answer the following two related questions: (i) what is achievable minimum rank (the optimal value of problem (\ref{mcomp-1})), and (ii) whether the minimum rank matrix, i.e., the optimal solutions to (\ref{mcomp-1}), is unique given a problem set-up. \yao{These result will also help to gain insights in the tradeoff in the theoretical properties of other matrix completion formulations, including LRMA and SDP formulations, compared with the original MRMC formulation.}

\yao{We show that given $m = |\Omega|$ observations  of an $n_1\times n_2$ matrix: (i) if the minimal rank $r^*$ is less than $\cR(n_1,n_2,m) := (n_1+n_2)/2 - [(n_1+n_2)^2/4 - m]^{1/2}$, then the corresponding solution is unstable: an arbitrary small perturbation of the observed values  can make this rank unattainable; (ii) if  $r^*>  \cR(n_1,n_2,m)$, then almost surely the solution is not (even locally) unique (cf., \cite{AShapiro_2017}).
This indicates that except on rare occasions, the MRMC problem cannot have both properties of possessing unique and stable solutions.
Consequently,  LRMA approaches (also used in \cite{DavenportRomberg16,Tropp17}) could be  a better alternative to the MRMC  formulation. Moreover, we argue  that the nuclear norm minimization approach is not (asymptotically) statistically  efficient (Section \ref{sec-semidef}).}


%


\subsection{Rank reducibility}

\label{sec-reducib}


We  denote by $r^*$ the optimal value of problem \eqref{mcomp-1}. That is, $r^*$ is the minimal rank of an $n_1\times n_2$ matrix with prescribed  elements    $M_{ij}$,  $(i,j)\in \O$. Clearly, $r^*$ depends on the index set $\O$ and values $M_{ij}$. A natural question is what values of $r^*$ can be attained. Recall that (\ref{mcomp-1}) is a non-convex problem and may  have multiple solutions. 

In a certain {\em generic sense}  it is possible to give a lower bound for the minimal rank $r^*$.  Let us  consider intersection of a set of low-rank matrices and the affine space of matrices satisfying the observation constraints.
Define the (affine) mapping $\A_M:\bbv_{\O^c}\to \bbr^{n_1\times n_2}$  as
\[
\A_M(X):= M+X,\;\;X\in \bbv_{\O^c}.
\]
As it has been pointed out before, the image $\A_M(\bbv_{\O^c})=M+\bbv_{\O^c}$ of mapping $\A_M$ defines the space of feasible points of the MRMC problem \eqref{mcomp-1}.
It is well known that $\M_r$ is a smooth, $C^\infty$,  manifold with
\begin{equation}\label{dimrank}
  \dim(\M_r)=r(n_1+n_2-r).
\end{equation}
It is said that  the mapping  $\A_M$
 intersects $\M_r$ {\em transverally}   if for every $X\in \bbv_{\O^c}$  either $\A_M(X)\not\in \M_r$, or  $\A_M(X)\in \M_r$
 and  the following  condition holds
\begin{equation}\label{transver}
 \bbv_{\O^c}+\T_{\M_r}(Y)=\bbr^{n_1\times n_2},
\end{equation}
 where $Y:=\A_M(X)$ and  $\T_{\M_r}(Y)$ denotes the tangent space to
$\M_r$ at $Y\in \M_r$  {\color{black} (we will give  explicit formulas for the tangent space $\T_{\M_r}(Y)$ in equations \eqref{tang-1} and \eqref{tan-1a} below.)}

  By using  a classical result of differential geometry,
 it is possible to show   that
  for {\em almost every}  (a.e.)
$M_{ij}$, $(i,j)\in \O$, the mapping   $\A_M$
 intersects $\M_r$   transverally (this holds for every     $r$) (see \cite{AShapiro_2017} for a discussion of this result). {\color{black} Transversality  condition
 \eqref{transver} means that the linear spaces $\bbv_{\O^c}$ and $\T_{\M_r}(Y)$ together span  the whole space $\bbr^{n_1\times n_2}$. Of course this cannot happen if the sum of their dimensions is less than  the dimension of $\bbr^{n_1\times n_2}$.}  Therefore  transversality  condition \eqref{transver}
   implies the following  dimensionality condition
 \begin{equation}\label{dimcond}
 \dim(\bbv_{\O^c})+\dim(\T_{\M_r}(Y))\ge \dim(\bbr^{n_1\times n_2}).
 \end{equation}
In turn the above condition \eqref{dimcond} can be written as
\begin{equation}\label{bound-3}
  r(n_1+n_2-r)\ge m,
\end{equation}
or equivalently $r\ge \cR(n_1,n_2,mm)$, where
\begin{equation}\label{bound-2}
\cR(n_1,n_2,m):=(n_1+n_2)/2-\sqrt{(n_1+n_2)^2/4- m}.
\end{equation}
 That is, if $r< \cR(n_1,n_2,m)$, then  the transversality  condition \eqref{transver} cannot hold and hence for a.e. $M_{ij}$ it follows that $\rank(M+X)\ne r$ for all $X\in \bbv_{\O^c}$.

 Now if $\A_M$
 intersects $\M_r$   transverally at $\A_M(X)\in \M_r$ (i.e., condition \eqref{transver} holds),
 then the intersection  $\A_M(\bbv_{\O^c})\cap \M_r$ forms a smooth manifold near the point $Y:=\A_M(X)$. When
 $r> \cR(n_1,n_2,m)$,  {\it this
  manifold has dimension greater than zero and hence the corresponding rank $r$ solution is not (locally) unique}.  This leads to the following  (for a  formal   discussion of these  results we can refer to  \cite{AShapiro_2017}).



\begin{theorem}[Generic lower bound and non-uniqueness of solutions]
\label{th-mrcm}
For any index set $\O$ of cardinality $m$ and   almost every
$M_{ij}$, $(i,j)\in \O$, the following  holds: {\rm (i)}   for every feasible point $Y$  of problem \eqref{mcomp-1} it follows that
\begin{equation}\label{bound-4}
\rank (Y)\ge \cR(n_1,n_2,m),
\end{equation}
{\rm (ii)} if   $r^*>\cR(n_1,n_2,m)$, then problem \eqref{mcomp-1} has multiple (more than one) optimal solutions.
\end{theorem}

It follows from part (i) of Theorem \ref{th-mrcm} that $r^*\ge \cR(n_1,n_2,m)$ for a.e. $M_{ij}$.
{\em Generically} (i.e., almost surely) the following lower bound for the  minimal rank $r^*$ holds
\begin{equation}\label{bound-5}
r^*\ge \cR(n_1,n_2,m),
\end{equation}
and  \eqref{mcomp-1} may have unique optimal solution only when $r^* = \cR(n_1,n_2,m)$.
Of course such equality could happen only if $\cR(n_1,n_2,m)$ is an integer number. As  Example \ref{ex-tight} below shows,   for any integer $ r^*\le \left\lceil \sqrt{m}\,\right\rceil$  satisfying \eqref{bound-5},  there exists an index set $\O$  such that the corresponding MRMC problem attains the minimal rank  $r^*$ for a.e. $M_{ij}$. In particular this shows that the lower bound   \eqref{bound-5} is tight.
When we have a square matrix  $n_1=n_2=n$, it follows that
\begin{equation}\label{bound-6}
\cR(n,n,m)=n-\sqrt{n^2 - m}.
\end{equation}
For $n_1=n_2=n$
and small $m/n^2$ we can approximate
\[
 \cR(n,n,m)=n\left(1-\sqrt{1-m/n^2}\right)
 \approx m/(2n).
\]
For example, for $n_1=n_2=1000$ and $m=20 000$ we have 
$\cR(n,n,m)=10.05$, and hence the bound
\eqref{bound-5} becomes
$r^*\ge 11$.


%
\begin{example}[Tightness of the  lower   bound  for $r^*$]
\label{ex-tight}
{\rm
For $r<\min\{n_1,n_2\}$ consider data matrix $M$ of the following form $M=\begin{psmallmatrix}
M_{1} &   0 \\ M_2 & M_{3}
\end{psmallmatrix}.
$
Here,  the three  sub-matrices $M_{1}$, $M_{2}$, $M_{3}$, of the respective order $r\times r$,   $(n_1 - r)\times r$ and  $(n_1 - r)\times (n_2-r)$, represent the observed entry values. Cardinality $m$ of the corresponding index set $\O$ is $r(n_1+n_2-r)$, i.e., here $r=\cR(n_1,n_2,m)$. Suppose that  the $r\times r$ matrix $M_{1}$ is nonsingular, i.e., its rows are linearly independent.
Then any row of matrix $M_2$ can be represented as a (unique) linear combination of rows of matrix $M_1$. It follows   that
the corresponding MRMC problem has (unique) solution of rank $r^*=r$.  In other words, the rank of the completed matrix will be equal to $r$ (the rank of the sub-matrix $M_1$) and there will be a unique matrix that achieves this rank.
Now suppose that some of the entries of the matrices $M_2$ and $M_3$ are not observed, and hence cardinality of the respective index set $\O$ is less than $r(n_1+n_2-r)$, and  thus  $r>\cR(n_1,n_2,m)$. In that case the respective minimal rank still is $r$, provided matrix $M_1$ is nonsingular, although the corresponding optimal solutions are not unique. In particular, if $M=\begin{psmallmatrix}
  M_{1} &   0\\0& 0\end{psmallmatrix}
$,
i.e., only the entries of matrix $M_1$ are observed, then $m=r^2$  and the minimum rank is $r$.
}
\end{example}

\subsection{Uniqueness of solutions of the  MRMC problem }
\label{sec-uniq}

\yao{Following Theorem \ref{th-mrcm},} for a given  matrix $M\in \bbv_\O$ and the corresponding minimal rank $r^*\le \cR(n_1,n_2,m)$, the question is whether the corresponding solution $Y^*$ of rank $r^*$ is unique.
 Although, the set of such  matrices $M$ is ``thin" (in the sense that it has Lebesgue measure zero),
 this question of uniqueness is important, in particular  for the statistical inference of rank (discussed in Section \ref{sec-stat}).
Available results, based on the so-called Restricted Isometry Property (RIP)  for low-rank matrix recovery from linear observations and based on the coherence property for low-rank matrix completion,  assert  that for certain probabilistic (Gaussian)  models such uniqueness holds with high  probability. However for a given matrix $M\in \bbv_\O$ it could be difficult  to verify whether the solution is unique   (some sufficient conditions for such uniqueness are given in \cite[Theorem 2]{Nowak16}, we will comment on this below.)

Let us consider the following concept of local uniqueness of  solutions.
\begin{definition}\label{def_local_uniqueness}
We say that an $n_1\times n_2$   matrix $\bar{Y}$ is a {\em locally} unique solution of problem \eqref{mcomp-1} if $P_\O(\bar{Y})=M$ and there is a neighborhood $\V\subset \bbr^{n_1\times n_2}$ of  $\bar{Y}$ such that $\rank (Y)\ne \rank(\bar{Y})$ for any $Y\in \V$, $Y\ne \bar{Y}$.
\end{definition}
Note that rank  is a lower semicontinuous function of matrix, i.e., if $\{Y_k\}$ is a sequence of matrices converging to matrix $Y$, then $\liminf_{k\to\infty} \rank(Y_k)\ge \rank (Y)$. Therefore local uniqueness of $\bar{Y}$ actually implies existence of the  neighborhood $\V$ such that $\rank (Y) > \rank(\bar{Y})$ for all $Y\in \V$, $Y\ne \bar{Y}$, i.e., that  at least locally problem \eqref{mcomp-1} does not have optimal solutions different from $\bar{Y}$.
The definition (\ref{def_local_uniqueness}) is closely related to the {\em finitely rank-$r$ completability} condition introduced in \cite{Nowak16}, which assumes that the MRMC problem has a finite number of rank $r$ solutions. Of course if problem \eqref{mcomp-1} has a non locally unique  solution of rank $r$, then the finitely rank-$r$ completability  condition cannot hold.

We now will introduce some constructions  associated with  the manifold $\M_r$ of  matrices of rank $r$. There are several  equivalent forms  how  the tangent space to the manifold $\M_r$ at $Y\in \M_r$ can be represented.  In one way it   can be written as
\begin{equation}\label{tang-1}
 \T_{\M_r}(Y)=\left\{Q_1 Y+Y Q_2: Q_1\in \bbr^{n_1\times n_1},\;
 Q_2\in \bbr^{n_2\times n_2}\right\}.
\end{equation}
In an  equivalent form  this  tangent space can be written as
\begin{equation}\label{tan-1a}
 \T_{\M_r}(Y)=\left\{H\in \bbr^{n_1\times n_2}:FHG=0  \right\},
\end{equation}
where $F$ is   an $(n_1-r)\times n_1$ matrix  of rank $n_1-r$
such that $F Y=0$ (referred to as
a {\em left side  complement} of $Y$) and $G$ is  an   $n_2\times (n_2-r)$ matrix   of rank $n_2-r$ such that    $Y G=0$ (referred to as a {\em right side  complement} of $Y$).
We also use the linear  space of matrices orthogonal (normal)  to $ \M_r$ at $Y\in \M_r$,   denoted by $\N_{\M_r}(Y)$.
A matrix  $Z$  is orthogonal to $ \M_r$ at $Y\in \M_r$ if and only if $\tr(Z^\top Y')=0$ for all $Y'\in \T_{\M_r}(Y)$. By \eqref{tang-1} this means that
\[
 \tr\left[Z^\top (Q_1 Y+Y Q_2)\right ]=0,\;\forall Q_1\in \bbr^{n_1\times n_1},\;\forall
 Q_2\in \bbr^{n_2\times n_2}.
\]
Since $ \tr\left[Z^\top (Q_1 Y+Y Q_2)\right ]=\tr (YZ^\top Q_1)+\tr (Z^\top Y Q_2)$ and matrices $Q_1$ and $Q_2$ are arbitrary, it follows that  the normal space can be written as
\begin{equation}\label{orthog-2}
\N_{\M_r}(Y)=\left\{Z\in\bbr^{n_1\times n_2}: Z^\top Y=0\;
{\rm and}\;
 Y Z^\top=0\right\}.
\end{equation}

\begin{definition}[Well-posedness condition]
\label{def-local}
We say that a matrix  $\bar{Y}\in \M_r$  is {\rm well-posed}, for problem  \eqref{mcomp-1},  if  $P_\O(\bar{Y})=M$ and the following condition holds
\begin{equation}\label{local-1}
 \bbv_{\O^c}\cap \T_{\M_r}(\bar{Y})=\{0\}.
\end{equation}
\end{definition}
%

%

\begin{figure}[h!]
\begin{center}
\includegraphics[width = 0.25\textwidth]{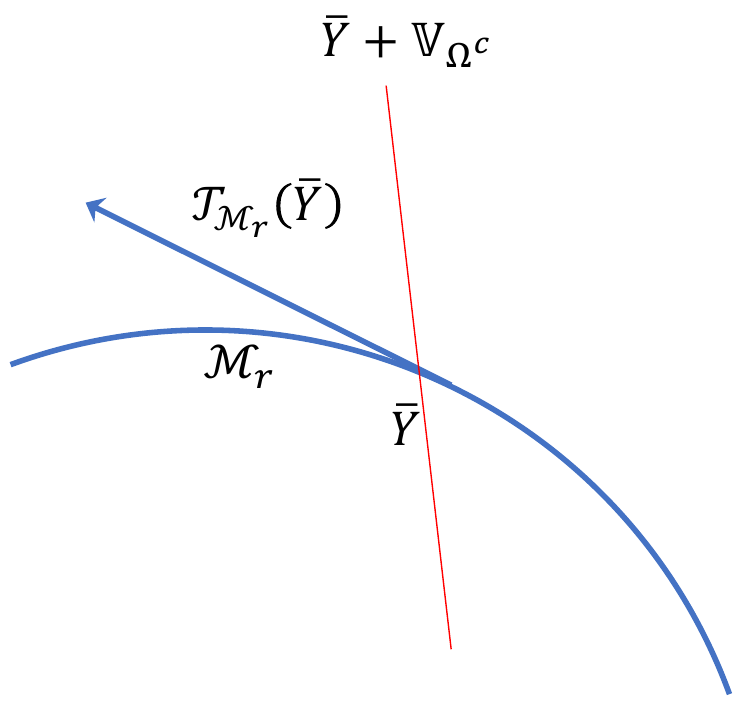}
\end{center}
\vspace{-0.2in}
\caption{Illustration of well-posedness condition.}
\label{fig:well-posed}
\end{figure}

Condition \eqref{local-1} (illustrated in Figure  \ref{fig:well-posed}) is a natural condition having a simple geometrical interpretation.  Intuitively, it means that the null space of the observation operator does not have any non-trivial matrix that lies in the tangent space of low-rank matrix manifold. Hence, there cannot be any local deviations  from the optimal solution that still satisfy the measurement constraints.
This motivates us to introduce the well-posedness condition that guarantees a matrix to be locally unique solution. \yao{Note that this is different from the so-called null space property \cite{Chandrasekaran2012} or the descent cone condition \cite{DavenportRomberg16}, which are for recovering sparse vectors, since the geometry therein is for sparse vectors whereas here we are dealing with manifold formed by low-rank matrices.}

Now we can give  sufficient conditions for local uniqueness:
\begin{theorem}[Sufficient conditions for local uniqueness]
\label{pr-uniq}
Matrix $\bar{Y}\in \M_r$ is a locally  unique solution of problem \eqref{mcomp-1} if $\bar Y$ is well-posed for \eqref{mcomp-1}.
\end{theorem}

\begin{remark}
{\rm
Suppose that   condition \eqref{local-1} does not hold, i.e., there exists nonzero matrix $H\in \bbv_{\O^c}\cap \T_{\M_r}(\bar{Y})$. This means that  there is a curve $Z(t)\in \M_r$ starting at $\bar{Y}$ and tangential to $H$, i.e., $Z(0)=\bar{Y}$ and $\|\bar{Y}+tH -Z(t)\|=o(t)$. Of course if moreover  $P_\O(Z(t))=M$ for all $t$ near $0\in \bbr$, then solution $\bar{Y}$ is not locally unique. Although this is not guaranteed, i.e., the sufficient condition \eqref{local-1} may be not necessary for local uniqueness of the solution  $\bar{Y}$, violation of this condition implies that solution $\bar{Y}$ is unstable in the sense that for some matrices $Y\in \M_r$ close to $\bar{Y}$ the distance $\|P_\O(Y)-M\|$ is of order $o(\|Y-\bar{Y}\|)$. {\color{black}  In that sense,   the well-posedness condition is   necessary  for local stability of solutions.}
}
\end{remark}


\subsection{\yao{Verifiable form of well-posedness condition}}

Below we present an equivalent form of the well-posedness condition \yao{that can be   verified  algebraically}.  By Theorem  \ref{pr-uniq} we have that if matrix  $\bar{Y}\in \M_r$ is well-posed, then $\bar{Y}$  is a locally  unique solution of problem \eqref{mcomp-1}.
Note that condition \eqref{local-1} implies that
$\dim(\bbv_{\O^c})+\dim(\T_{\M_r}(\bar{Y}))\le n_1 n_2$. That is, condition \eqref{local-1} implies that $r(n_1+n_2-r)\le m$ or equivalently  $r\le \cR(n_1,n_2,m)$.
By Theorem \ref{th-mrcm} we have that if $r^*> \cR(n_1,n_2,m)$, then the corresponding optimal solution cannot be   locally  unique almost surely.
Note  that since the space $\bbv_{\O}$ is orthogonal to the space $\bbv_{\O^c}$,
by duality arguments condition \eqref{local-1} is equivalent to the following condition
\begin{equation}\label{local-2}
 \bbv_{\O} +  \N_{\M_r}(\bar{Y})=\bbr^{n_1\times n_2}.
\end{equation}
By using formula \eqref{tan-1a} it is also possible to write condition \eqref{local-1} in the following  form
\begin{equation}\label{loc-1a}
 \{X\in  \bbv_{\O^c}:FXG=0\}=\{0\},
\end{equation}
where $F$ is  a left
side  complement of $\bar{Y}$ and $G$ is a right
side  complement of $\bar{Y}$.
Recall  that
$
\vect(FXG)=(G^\top\otimes F)\vect (X).
$
Column vector of matrix $G^\top\otimes F$ corresponding to component   $x_{ij}$  of vector $\vect (X)$,  is $g_{j}^\top \otimes  f_i$, where $f_i$ is the $i$-th column of matrix $F$ and $g_j$ is the $j$-th row of matrix $G$.
Condition \eqref{loc-1a} means that the column vectors $g_{j}^\top \otimes  f_i$, $(i,j)\in \O^c$,   are linearly independent. {\color{black} Then we obtain the following
verifiable condition for  checking the well-posedness of a given solution:}

\begin{theorem}[Equivalent condition  of well-posedness]
\label{pr-uniqa}
Matrix $\bar{Y}\in \M_r$ satisfies condition \eqref{local-1}  if and only if  for any   left side  complement $F$   and  right
side  complement $G$  of $\bar{Y}$,
the column vectors $g_{j}^\top \otimes  f_i$, $(i,j)\in \O^c$,   are linearly independent.
\end{theorem}
A consequence of the theorem is that if $\bar{Y}\in \M_r$ is  well-posed, then necessarily
$(n_1-r)(n_2-r)\ge |\O^c|$, since vectors $g_{j}^\top \otimes  f_i$ have dimension $(n_1-r)(n_2-r)$. Since $|\O^c|=n_1n_2-m$, this is equivalent to $r(n_1+n_2-r)\le m$. That is, the well-posedness cannot happen if $r>\cR(n_1,n_2,m)$. This of course is not surprising in view of discussion of Section \ref{sec-reducib}.

Theorem \ref{pr-uniqa} also implies the following necessary condition for well-posedness of $\bar{Y}\in \M_r$ in terms of the pattern of the index set  $\Omega$, \yao{which is related to the completability condition in \cite{Nowak16} that each row and each column has at least $r$ observations}.
If  matrix $\bar{Y}\in \M_r$ is  well-posed  for problem  \eqref{mcomp-1}, then at each row and each column of $\bar{Y}$ there are at least $r$ elements of the index set $\O$.
Indeed,  suppose that in row $i\in \{1,...,n_1\}$ there are less than $r$ elements of $\O$. This means that the set $\sigma_i:=\{j: (i,j)\in \O^c\}$ has cardinality greater than $n_2-r$. Let $F$ be  a left side  complement of
$\bar{Y}$ and  $G$ be  a right side   complement of
$\bar{Y}$. Since rows $g_j$ of $G$ are of dimension $1\times (n_2-r)$, we have then that vectors $g_j$, $j\in \sigma_i$, are linearly dependent, i.e., $\sum_{j\in \sigma_i}\lambda_j g_j=0$ for some $\lambda_j$, not all of them zero. Then
\begin{equation}
\begin{array}{l}
\sum_{j\in \sigma_i}\lambda_j (g_{j}^\top \otimes  f_i)=
\big(\sum_{j\in \sigma_i}\lambda_j  g_{j}\big)^\top \otimes f_i=0.
\end{array}
\label{condition_unique}
 \end{equation}
This contradicts the condition for vectors $g_{j}^\top \otimes  f_i$, $(i,j)\in \O^c$,  to be  linearly independent.
Similar arguments can be applied to the columns of matrix
$\bar{Y}$.
This necessary condition for well-posedness  is not surprising since if there is a row with less than
$r$ elements of $\O$, then this row in not uniquely defined in the corresponding rank $r$ solution (cf., \cite{Nowak16}).
 However, although necessary, the condition for the index set  $\O$ to have at each  row and each column at least $r$ elements is not sufficient to ensure well-posedness as shown by
 Theorem  \ref{pr-reduc}  below.  Note that by definition the matrices  $F$ and $G$ are of full rank.

\subsection{Generic nature of the well-posedness}
In a certain sense the well-posedness condition is generic, as we explain below.
Denote by   $\F_r\subset \bbr^{n_1\times r}$ and $\G_r\subset \bbr^{n_2\times r}$ the respective sets  of matrices of rank $r$.  Consider the set $\Theta:=\F_r\times \G_r\times \bbv_{\O^c}$ viewed as a subset of $\bbr^{n_1 r+n_2 r+n_1n_2-m}$,
 and  mapping
$\cF:\Theta\to \bbr^{n_1\times n_2}$ defined as \[\cF(\theta):=V W^\top+X,\;\;\theta=(V,W,X)\in \Theta.
\]
Note that the sets $\G_r$ and
$\F_r$ are open and connected, and hence the set $\Theta$ is open and connected, and the components of  mapping $\cF(\cdot)$ are  polynomial  functions.

Let $\Delta(\theta)$ be the Jacobian of mapping $\cF$. That is, $\Delta(\theta)$ is $(n_1 r+n_2 r+n_1n_2-m)\times (n_1 n_2)$ matrix of partial derivatives of $\cF(\theta)$ taken  with respect to  a specified order of the   components of the corresponding  matrices. Let us consider the following concept  associated with rank $r$ and index set $\O$  (cf., \cite{sha1986}).
\begin{definition}
\label{def-chr}
We refer to
\begin{equation}\label{chrank}
\varrho:=\max_{\theta\in \Theta}\left\{\rank\big (\Delta(\theta)\big)\right\}
\end{equation}
as the  {\em characteristic  rank} of mapping $\cF$ and say that   $\theta\in \Theta$ is a {\em  regular point } of $\cF$ if $\rank\big (\Delta(\theta)\big)=\varrho$. We say that $(V,W)\in \F_r\times \G_r$ is regular if  $\theta=(V,W,X)$ is regular for some  $X\in  \bbv_{\O^c}$.
\end{definition}

Since $\cF(V,W,\cdot)$ is linear, the Jacobian  $\Delta(V,W,X)$ is the same for all $X\in \bbv_{\O^c}$, i.e.,
$\Delta(V,W,X)= \Delta(V,W,X')$ for any $X,X'\in  \bbv_{\O^c}$ and
$(V,W)\in \F_r\times \G_r$. Hence  if a point  $\theta=(V,W,X)$ is regular for some  $X\in  \bbv_{\O^c}$, then  $(V,W,X')$ is regular for any  $X'\in  \bbv_{\O^c}$. Therefore regularity actually is a property of points $(V,W)\in \F_r\times \G_r$.
Since
$V W^\top\in \M_r$ for $(V,W)\in \F_r\times \G_r$,  and the dimension of manifold $\M_r$ is $r(n_1+n_2-r)$ it follows that $\varrho\le \cf(r,m)$ where
\begin{equation}\label{rankineq}
\cf(r,m):= r(n_1+n_2-r)+n_1n_2-m.
\end{equation}

\begin{theorem}
\label{th-wpgen}
The following holds. (i) Almost every point  $(V,W)\in \F_r\times \G_r$ is regular. (ii) The set of regular points forms an open subset of $\F_r\times \G_r$. (iii)  For any regular point   $(V,W)\in\F_r\times \G_r$, the corresponding  matrix   $Y=V W^\top$ satisfies the well-posedness condition \eqref{local-1}
if and only if  the characteristic rank $\varrho$ is equal to $\cf(r,m)$. (iv)  If $\varrho<\cf(r,m)$ and a point   $(\bar{V},\bar{W})\in\F_r\times \G_r$ is regular, then for any $Y\in\M_r$ in a neighborhood of $\bar{Y}=\bar{V}\bar{W}^\top$  there exists $X\in \bbv_{\O^c}$ such that $Y=\bar{Y} +X$.
\end{theorem}

The significance of Theorem \ref{th-wpgen} is that this shows that for given rank $r$ and index set $\O$, either $\varrho=\cf(r,m)$ in which case
a.e. $Y\in \M_r$ satisfies the well-posedness condition \eqref{local-1},  or $\varrho<\cf(r,m)$ in which case condition \eqref{local-1}  does not hold for all $Y\in \M_r$ and generically rank $r$ solutions are not locally unique.

We have that  a necessary condition for  $\varrho=\cf(r,m)$  is that each row and each column of the considered matrix has at least $r$ observed entries. Another necessary condition is for the index set to be irreducible (see Theorem \ref{pr-reduc}). Whether these two conditions are sufficient for $\varrho=\cf(r,m)$ to hold remains an open question, but numerical experiments, reported in Section \ref{sec-numer},  indicate that in a certain probabilistic sense  chances of occurring  not well posed solution are negligible when $r$ is slightly less than $\cR(n_1,n_2,m)$.


\subsection{Global uniqueness of solutions for special cases}
In some rather special   cases it is possible to give  verifiable conditions for  global uniqueness of minimum rank solutions.   The following  conditions are straightforward extensions of    well known conditions  in Factor Analysis (cf., \cite[Theorem 5.1]{andrub} ).

\begin{assu}
\label{assump-1}
Suppose that: {\rm (i)} for a given  index  $(k,l) \in \O^c$,
  there exist index sets  ${\cal I}_1\subset \{1,...,n_1\}\setminus \{k\}$ and ${\cal I}_2 \subset \{1,...,n_2\}\setminus \{l\}$ such that $|{\cal I}_1|=|{\cal I}_2|=r$,   ${\cal I}_1\times {\cal I}_2\subset \O$, and  $\{k\}\times {\cal I}_2\subset \O$ and $\{l\}\times {\cal I}_1\subset \O$, {\rm (ii)} the $r\times r$ submatrix of $M$ corresponding to rows $i\in{\cal I}_1$ and columns $j\in {\cal I}_2$  is nonsingular.
\end{assu}

For example, for $r=1$ part (i) of the above assumption means existence of indexes $k'\ne k$ and $l'\ne l$ such that $(k',l),(k,l'),(k',l')\in \O$.

\begin{proposition}
\label{pr-gluniq}
Suppose that Assumption \ref{assump-1} holds for an  index  $(k,l) \in \O^c$.
Then  the minimum rank $r^*\ge r$, and
for any matrix $Y\in \M_r$ such that $P_\O(Y)=M$ it follows  that $Y_{kl}=\bar{Y}_{kl}$.
\end{proposition}

Clearly  part  (ii) of   Assumption \ref{assump-1} implies that  $r^*\ge r$.
The other   result of the above proposition  follows by observing that the $(r+1)\times (r+1)$ submatrix of $Y$ corresponding to rows $\{k\}\cup  {\cal I}_1$ and columns $\{l\}\cup {\cal I}_2$ has rank $r$ and hence zero determinant, and applying Shur complement for the element $Y_{kl}$.  {\color{black}  Note that provided the part (i) holds,  part  (ii) is generic in the sense that it holds for a.e. $M_{ij}$.

 If Assumption \ref{assump-1}  holds for {\em  every} $(k,l) \in \O^c$, then the  uniqueness of the solution $\bar{Y}$ follows. This is   closely related to \cite[Theorem 2]{Nowak16}, but is not the same. It is assumed in  \cite{Nowak16} that every column of $M$ has $r+1$ observed entries. For example, consider $2\times 2$ matrix with 3  observed entries, $M_{12}= M_{21}= M_{22}=1$. The only unobserved entry, corresponding to the index $(1,1)$, satisfies Assumption \ref{assump-1} and rank one matrix, with all entries equal 1, is the unique solution of the MRMC problem. On the other hand the first column of matrix $M$ has only one observed entry.}

\begin{remark}
\label{rem-wilson}
{\rm \yao{This result has been observed in an much earlier paper by } Wilson  and Worcester \cite{wilson39}, where an example was constructed  of two   $6\times 6$ symmetric matrices of rank 3 with the same off-diagonal  and different diagonal elements.  If we define the index set as  $\O:=\{(i,j): i\ne j,\;i,j=1,...,6\}$, then this can be viewed as an example of two different locally unique solutions of rank 3. Note that here $m=30$ and  $\cR(6,6,30)=6-\sqrt{6}$. That is $\cR(6,6,30)>3$ and generically (almost surely)  rank cannot be reduced below $r=4$. We will discuss this example further in Section \ref{sec-numer}. }
\end{remark}

\subsection{Identifiable $\Omega$}

\yao{Our results can also be used  to determine wether observation patterns $\Omega$ is identifiable.}
First note that uniqueness of the minimum rank solution is invariant with respect to permutations of  rows and columns of  matrix $M$. This motivates to introduce the following definition.

 {\color{black}
 \begin{definition}
\label{def-reduc}
 We say that the index set $\O$ is {\em reducible} if by permutations of rows and columns,  the set $\O$ can be represented as  the  union $\O'\cup \O''$ of two disjoined sets $\O'\subset\{1,...,k\}\times \{1,...,l\}$ and $\O''\subset\{k+1,...,n_1\}\times \{l+1,...,n_2\}$ for some $1\le k<n_1$ and $1\le l<n_2$.
Otherwise we say that $\O$ is irreducible.
\end{definition}

 Reducibility of   the index set $\O$ means that   by permutations of rows and columns,  matrix $M$ can be represented in the  block diagonal form
\begin{equation}
M=\left[
  \begin{array}{ccc}
  M' &    0 \\
 0 & M''
  \end{array}
  \right],
  \label{M_block}
\end{equation}
where matrices  $M'$ and $M''$ are of order $k\times l$ and $(n_1-k)\times (n_2-l)$, respectively, with  observed entries  $M'_{ij}$, $(i,j)\in \O'$,  and
 $M''_{ij}$, $(i,j)\in \O''$.
Some entries of matrices $M'$ and $M''$ can also be zero if the corresponding entries of matrix $M$ are zeros.
}

\begin{theorem}[Reducible index set]
\label{pr-reduc}
If the index set $\O$ is   reducible, then any minimum rank solution $\bar{Y}$   
is not locally (and hence globally) unique.
\end{theorem}

As it was shown in Theorem \ref{pr-uniq}, if $\bar{Y}$ is not locally unique, then it cannot be well-posed. Therefore if the index set $\O$ is reducible, then any minimum rank solution    is not well-posed. Of course even if $\O$ is reducible, it still  can happen that in each row and column there are at least $r$ elements of the index set $\O$.
That is, the condition of having $r$ elements of the index set $\O$ in each row and column
is not sufficient to ensure the well-posedness property.

{\color{black}
\begin{remark}
{\rm
Reducibility/irreducibility of the index set $\O$  can be verified in the following way. Consider the undirected graph $G=(V,E)$ with the set of  vertices $V:=\O$, and edges between two vertices $(i,j),(i',j')\in \O$ iff $i=i'$ or $j=j'$.
Then $\O$ is irreducible iff $G$ has only one connected component. A connected component  of $G$  is a subgraph in which any two vertices are connected to each other by paths, and which is connected to no additional vertices in the supergraph $G$. There are algorithms of running time $O(|V|+|E|)$ which can find every vertex that is
reachable from a given vertex of $G$, and hence   to determine a connected   component  of $G$,
e.g., the well known {\em breadth-first search}
algorithm  \cite[Section 22.2]{CLRS}.     Note that the number of vertices in $G$  is $m=|\O|$, which could be much smaller than $n_1n_2$.    }
\end{remark}
}





\subsection{Uniqueness of rank one solutions}

In this section we discuss uniqueness of rank one solutions of the MRMC problem \eqref{mcomp-1}. We show that in  case of the minimum rank one, irreducibility of $\O$ is sufficient for the global uniqueness.
We assume that all $M_{ij}\ne 0$, $(i,j)\in \O$, and that every row and every column of the matrix $M$ has at least one element $M_{ij}$. Let $\bar{Y}$ be a solution of rank one of  problem \eqref{mcomp-1}, i.e., there are  nonzero column  vectors $v$ and $w$ such that
$\bar{Y}=v w^\top$ with $P_\O(\bar{Y})=M$.

Recall that permutations of the components of vector $v$ corresponds to permutations of the rows of the respective rank one matrix, and permutations of the components of vector $w$ corresponds to permutations of the columns  of the respective rank one matrix. It was shown in Theorem \ref{pr-reduc} that if the index set $\O$ is reducible, then solution $\bar{Y}$ cannot be locally unique. In case of rank one solution the converse of that also holds.

\begin{theorem}[Global uniqueness for rank one solution]
\label{pr-rankone}
Suppose that $\O$ is irreducible,  $M_{ij}\ne 0$ for all $(i,j)\in \O$, and   every row and every column of the matrix $M$ has  at least one element $M_{ij}$, $(i,j)\in \O$. Then   any  rank one solution is globally unique.
\end{theorem}

{\color{black}
It could be mentioned that even for $r=1$ the  irreducibility is a weaker condition than part (i) of Assumption \ref{assump-1}  applied to every $(k,l) \in \O^c$. For example, let $n_1=n_2=n\ge 3$ and $\O=\{(i,j):i\ge j,\;i,j=1,...,n\}\setminus\{(n,1)\}$.
This set $\O$ irreducible. However for the index $(1,n)$, Assumption \ref{assump-1}(i) does not hold.
 }

\subsection{Semidefinite relaxations
}
\label{sec-semidef}


%

\yao{Consider  the trace minimization problem \eqref{sim-3} (which can be viewed as a generalized version of the nuclear norm minimization problem)}, and  assume that the  matrix $C\in \bbw_{\cS^c}$ is {\em positive definite}.
The  (Lagrangian) dual of problem \eqref{sim-3} is the problem
 \begin{equation}\label{comp-11}
 \max_{\Lambda\succeq 0}\min_{X\in \bbw_{\cS^c}} \tr(C  X) -\tr[\Lambda (\Xi+X)].
\end{equation}
For $\Lambda=C-\Theta$, with  $\Theta \in\bbw_{\cS}$,   problem \eqref{comp-11} can be written  (note that  $\tr(C  \Xi)=0$ for $\Xi\in \bbw_{\cS}$) as
\begin{equation}\label{comp-12}
 \max_{\Theta\in \bbw_{\cS}} \tr( \Theta  \Xi)\;{\rm subject\; to}\; C-\Theta\succeq 0.
\end{equation}



We have the following uniqueness  results for the SDP approach, which is a consequence of (cf., \cite[Theorem 5.2] {sha85} and \cite[Proposition 8]{AShapiro_2017}) (we also provide justification in the appendix):
\begin{theorem}
\label{pr-sdpuniq}
{\rm (i)}
 For a given $\Xi\in \bbw_{\cS}$ it follows that for
almost every   positive definite matrix $C\in \bbw_{\cS^c}$,  problem   \eqref{sim-3}  has unique optimal solution.
{\rm (ii)} For a given positive definite matrix $C\in \bbw_{\cS^c}$ it follows that for  almost every $\Xi\in \bbw_{\cS}$ the dual problem \eqref{comp-12} has unique optimal solution.
\end{theorem}

\yao{However, we have the following observation, which comes as a consequence of \cite[Theorem 2]{AShapiro_2017}:}
\begin{remark}
\label{rem-as2}
{\rm
\alex{
Consider the minimum trace (MT)  problem \eqref{sim-2}. Suppose that the matrix $\Xi$ is observed with errors: $\Xi=\Xi^*+U$, where     $U\in \bbs^p$ is  random matrix such that $N^{1/2}U$ converges in distribution to a random matrix $\Upsilon\in \bbs^p$ whose entries have zero means and finite positive second order moments (we discuss a similar model for the MRMC in section \ref{sec-approx} below).
Let  $\hat{X}$ and $X^*$ be optimal solutions of the  MT problems of the form  \eqref{sim-2}  for matrices $\Xi$ and $\Xi^*$, respectively. Then under mild regularity conditions
\begin{equation}\label{astr}
 \tr(\hat{X})-\tr(X^*)=\sup_{\Lambda\in \sol(D)}\tr (\Lambda U) +o_p(N^{-1/2}),
\end{equation}
where $\sol(D)$ is the set of optimal solutions of the dual problem \eqref{comp-11}.
 When the minimal rank of the true model is less than the generic lower bound (given by the right hand side of \eqref{mrfa}), the set  $\sol(D)$ contains  more than one element. Consequently  $\tr(\hat{X})$,  considered as an estimator of $\tr(X^*)$, has a bias of order $N^{-1/2}\bbe\big[\sup_{\Lambda\in \sol(D)}\tr (\Lambda \Upsilon)\big]$  (we can refer to  \cite[Theorem 2]{AShapiro_2017}   for derivations and a discussion  of the required regularity conditions).
}}
\end{remark}

We conclude this section by mentioning
  connections  to existing results in Factor Analysis.
The classical Minimum Rank Factor Analysis (MRFA) can be viewed as a particular case of problem \eqref{sim-1} with   $\bbw_{\cS^c}$ being the space $\bbd^p$ of $p\times p$ diagonal matrices, and given  symmetric  matrix $\Xi$ of off diagonal elements. It is possible to show that generically (i.e., for a.e. $\Xi$) the reduced rank   of the MRFA problem is bounded  (cf., \cite{shap1982}):
\begin{equation}\label{mrfa}
 \rank(\Xi+X)\ge \frac{2p+1-\sqrt{8p+1}}{2},\;\forall X\in \bbd^p.
\end{equation}
In Factor Analysis the respective minimum trace problem of the form \eqref{sim-2} is called the Minimum Trace Factor Analysis (MTFA). A relation between MRFA and MTFA problems is discussed in \cite{shap1982,shap82b}. In Factor Analysis conditions analogues to the assumptions of  Proposition  \ref{pr-gluniq} can be used to show that in a certain generic sense,  MRFA  solution is unique if the respective minimal rank is less than $p/2$ (we can refer to \cite{bekker}, and references therein, for a discussion of uniqueness of MRFA solutions).

\subsection{LRMA and its properties}
\label{sec-approx}


 We discuss below the LRMA approach \eqref{leastsq}. 
%
%
Compared with the formulation of exact low rank recovery,  the LRMA is more realistic  in the presence of noise.
By Theorem \ref{th-mrcm} we have that if the minimal rank $r^*$ is less than $\cR(n_1,n_2,m)$, then the corresponding solution is unstable in the sense that an arbitrary small perturbation of the observed values  $M_{ij}$ can make this rank unattainable.
On the other hand if  $r^*>  \cR(n_1,n_2,m)$, then almost surely the solution is not (even locally) unique.
This  indicates that except in  rare occasions, problem \eqref{mcomp-1}  of exact rank minimization   cannot have both properties of  possessing   unique and stable  solutions.
Consequently,  what makes sense is to try  to solve the minimum rank problem approximately.

\begin{proposition}[Necessary condition for LRMA]
\label{pr-opt}
The following are necessary conditions for $Y\in \M_r$ to be an optimal solution of problem \eqref{leastsq}
\begin{equation}
\label{optnes-1}
 (P_\O(Y)-M)^\top Y=0\;\;{\rm and}\;\;
 Y (P_\O(Y)-M)^\top=0.
\end{equation}
\end{proposition}

\begin{remark}
\label{rem-uniq}
{\rm
We can view the least squares problem (\ref{leastsq}) from the following point of view. Consider   function
\begin{equation}\label{func-g}
 \phi(Y,\Theta):=\half \tr[(P_\O(Y)-\Theta)^\top(P_\O(Y)-\Theta)],
\end{equation}
with $\Theta\in \bbv_\O$ viewed as a parameter. Define
\begin{equation}\label{func-f}
\begin{split}
f(Y)&:=\half \sum_{(i,j)\in \O}\left(Y_{ij}-M_{ij}\right)^2\\
&=\half \tr[(P_\O(Y)-M)^\top(P_\O(Y)-M)],
\end{split}
\end{equation}
Hence, the problem \eqref{leastsq} consists of minimization of $f(Y)$ subject to $Y\in \M_r$.
Note that for $\Theta=M$ we have  $f(\cdot)=\phi(\cdot,M)$, where $f(\cdot)$ is defined in \eqref{func-f}.
Let   $\bar{Y}\in \M_r$ be such that $\phi(\bar{Y},\Theta_0)=0$ for some $\Theta_0\in \bbv_\O$,   i.e., $P_\O(\bar{Y})=\Theta_0$.
A sufficient condition for $\bar{Y}$ to be  a   locally unique solution of problem \eqref{mcomp-1},  at $M=\Theta_0$, is
\begin{equation}\label{sufcon}
\tr\left[P_\O(H)^\top P_\O(H)\right]>0,\;\;\forall H\in \T_{\M_r}(\bar{Y})\setminus\{0\}.
\end{equation}
The above condition   means that if $H\in \T_{\M_r}(\bar{Y})$  and $H\ne 0$,   then $P_\O(H)\ne 0$.
In other words this means that  the kernel  $${\rm Ker}(P_\O):= \{H\in \T_{\M_r}(\bar{Y}):P_\O(H)=0\}$$  is $\{0\}$.
Since $P_\O(H)= 0$ for any $H\in \bbv_{\O^c}$, it follows  that:    {\em condition \eqref{sufcon} is equivalent to  the sufficient condition \eqref{local-1} of Proposition} \ref{pr-uniq}.  That is, condition \eqref{sufcon}  means that matrix $\bar{Y}$ is well-posed for problem \eqref{mcomp-1}.
}
\end{remark}

Assuming that condition  \eqref{sufcon} (or equivalently condition \eqref{local-1}) holds, by applying the Implicit Function Theorem to the first order optimality conditions of the least squares problem (\ref{leastsq}) we have  the following result.

\begin{proposition}
\label{pr-locuniq}
Let $\bar{Y}\in \M_r$ be such that $P_\O(\bar{Y})=\Theta_0$ for some $\Theta_0\in \bbv_\O$ and suppose that the well posedness  condition
\eqref{local-1} holds. Then there exist neighborhoods $\V$ and $\W$ of $\bar{Y}$ and $\Theta_0$, respectively, such that for any $M\in \W\cap \bbv_\O$  there exists unique $Y\in \V\cap \M_r$ satisfying the optimality conditions \eqref{optnes-1}. \end{proposition}

The above proposition implies the following. Suppose that we run a numerical procedure which identifies a matrix $\bar{Y}\in \M_r$ satisfying the (necessary) first order   optimality conditions \eqref{optnes-1}. Then if $P_\O(\bar{Y})$ is sufficiently close to $M$  (i.e., the fit $\sum_{(i,j)\in \O}\left(Y_{ij}-M_{ij}\right)^2$ is sufficiently small)  and condition  \eqref{local-1} holds at $\bar{Y}$,  then we can say that $f(Y)>f(\bar{Y})$ for all $Y\ne \bar{Y}$ in a neighborhood of $\bar{Y}$. That is, $\bar{Y}$ solves the least squares  problem at least locally. Unfortunately it is not clear how to quantify the ``sufficiently close" condition, and this does not guarantee global optimality of $\bar{Y}$ unless $\bar{Y}$ is the unique minimum rank solution.

\section{Statistical test for rank selection}
\label{sec-stat}

\yao{In this section, we propose a  statistical test procedure for  value of the ``true'' minimal rank,  when the  entries of the data matrix  $M$  are observed with noise. Such statistical approach can be useful for many existing low-rank matrix completion algorithms, which  require a pre-specification of the matrix rank, such as the alternating minimization approach to solving the non-convex problem by representing the low-rank matrix as a product of two low-rank matrix factors (see, e.g.,  \cite{DavenportRomberg16}).}

Consider this for the LRMA formulation. By the above discussion, it will be natural to take some value of $r$ less than $\cR(n_1,n_2,m)$,  since otherwise we will not even have locally unique solution. Can the fit of $Y\in \M_r$ to $X+M$, and hence the choice of $r$, be tested in some statistical sense?

To proceed we assume the following model with noisy and possibly biased observations of a subset of matrix entries. There is a (population) value $Y^*$ of $n_1\times n_2$ matrix of rank $r < \cR(n_1,n_2,m)$ and $M_{ij}$  are viewed as observed (estimated) values of $Y^*_{ij}$, $(i,j)\in \O$, based on a sample of size $N$.
The observed values are modeled as
\begin{equation}\label{observ}
M_{ij}=Y^*_{ij}+N^{-1/2} \Delta_{ij}+\e_{ij}, \;(i,j)\in \O,
 \end{equation}
where $Y^*\in \M_r$ and    $\Delta_{ij}$ are some (deterministic) numbers.  The random   errors $\e_{ij}$ are assumed to be    independent of each other and such that $N^{1/2}\e_{ij}$ converge  in distribution to normal with mean zero and variance $\sigma^2_{ij}$, $(i,j)\in \O$.
The additional terms $N^{-1/2} \Delta_{ij}$ in \eqref{observ} represent  a possible deviation of population values from the ``true'' model and are often referred to as the population drift or a sequence of local alternatives (we can refer to \cite{McManus} for a historical overview of invention of  the local alternatives setting). This is a reasonably  realistic model motivated by many real applications.

 \begin{definition}
  We say that the model is {\em globally  identifiable}   (at $Y^*$)  if $\bar{Y}\in  \bbr^{n_1\times n_2}$ of $\rank(\bar{Y})\le r$ and  $P_\O(\bar{Y})=P_\O(Y^*)$ imply that $\bar{Y}=Y^*$, i.e., $Y^*$ is the unique   solution of the respective matrix completion problem. Similarly it is said that  the model is {\em locally  identifiable} if this holds for all such $\bar{Y}$ in a neighborhood of   $Y^*$, i.e.,
$Y^*$ is a locally unique solution.
\end{definition}

Consider the following weighted least squares problem (\yao{a generalization of (\ref{leastsq})}):
\begin{equation}\label{weight}
   \min_{Y\in \M_r}\sum_{(i,j)\in \O}w_{ij}\left(M_{ij}-Y_{ij} \right)^2,
\end{equation}
for some weights  $w_{ij}>0$, $(i,j)\in \O$. (Of course, if $w_{ij}=1$, $(i,j)\in \O$, then problem \eqref{weight} coincides with the least squares problem \eqref{leastsq}.)
We have the following standard result about  consistency of the least squares estimates.

\begin{proposition}
\label{pr-cons}
Suppose that the model is globally identifiable at $Y^*\in \M_r$ and values $M_{ij}$, $(i,j)\in \O$,  converge in probability to the respective values $Y^*_{ij}$ as the sample size $N$ tends to infinity. Then an optimal solution $\hat{Y}$ of problem \eqref{weight} converges in probability to $Y^*$  as $N \rightarrow \infty$.
\end{proposition}


 Consider the following weighted least squares test statistic
\begin{equation}\label{stat-1}
 T_N(r):=N \min_{Y\in \M_r}\sum_{(i,j)\in \O}w_{ij}\left(M_{ij}-Y_{ij}\right)^2,
\end{equation}
where $w_{ij}:=1/\hat{\sigma}^{2}_{ij}$ with $\hat{\sigma}^{2}_{ij}$ being consistent estimates of $\sigma^{2}_{ij}$ (i.e.,  $\hat{\sigma}^{2}_{ij}$ converge  in probability to $\sigma^{2}_{ij}$ as $N\to\infty$).
Recall that the respective  condition of form \eqref{local-1}, or equivalently    \eqref{sufcon}, is sufficient for local identifiability of $Y^*$.
The following asymptotic  results can be compared with similar results in the analysis of covariance structures   (cf., \cite{ste1985}).

\begin{proposition}[Asymptotic properties of test statistic]
\label{th-test}
 Consider  the noisy observation model  \eqref{observ}. Suppose that  the model is globally identifiable at $Y^*\in \M_r$   and $Y^*$ is
well-posed  for problem  \eqref{mcomp-1}. Then as $N\to\infty$,    the test statistic $T_N(r)$ converges in distribution to noncentral $\chi^2$ distribution with degrees of freedom ${\rm df}_r =m-r(n_1+n_2-r)$ and the  noncentrality parameter
\begin{equation}\label{noncen}
\delta_r= \min\limits_{H\in  \T_{\M_r}(Y^*) }\sum_{(i,j)\in \O}\sigma^{-2}_{ij}\left(\Delta_{ij}-H_{ij}\right)^2.
\end{equation}
\end{proposition}

Note that the optimal (minimal) value of the weighted least squares  problem \eqref{weight} can be approximated by
\begin{equation}\label{stat-2}
\min_{H\in  \T_{\M_r}(Y^*)}\sum_{(i,j)\in \O}w_{ij}\left(E_{ij}-H_{ij}\right)^2+R_N,
\end{equation}
with
$E_{ij}:=N^{-1/2} \Delta_{ij}+\e_{ij}$ and the error term                                 $R_N=o\left (\|M-P_\O(Y^*)\|^2\right )$ being of stochastic order $R_N=o_p(N^{-1})$.
Hence, the noncentrality parameter, given in \eqref{noncen}, can be approximated as
\begin{equation}\label{nonc-2}
\delta_r\approx N \min_{Y\in \M_r}\sum_{(i,j)\in \O}w_{ij}\left(Y^*_{ij}+N^{-1/2} \Delta_{ij} -Y_{ij} \right)^2. \end{equation}
That is, the noncentrality parameter is approximately equal to $N$ times the fit to the ``true'' model  of the alternative  population values
$Y^*_{ij}+N^{-1/2} \Delta_{ij}$ under small perturbations of order  $O(N^{-1/2})$.

\begin{remark}
\label{rem-model}
{\rm
The above asymptotic results are formulated in terms of the ``sample size $N$" suggesting that the observed values are estimated from some data.
{\color{black} That is, the given  values $\bar{M}_{ij}$,  $(i,j)\in \O$,  are obtained by averaging   i.i.d.  data points $M_{ij}^\ell$, $\ell=1,...,N$. In that case asymptotic normality of $N^{1/2}\e_{ij}$ can be justified by application of the Central Limit Theorem, and the corresponding variances $\sigma^2_{ij}$ can be estimated from the data in the usual way
$\hat{\sigma}^2_{ij}=(N-1)^{-1}\sum_{\ell=1}^N(M_{ij}^\ell-\bar{M}_{ij})^2$.}
This model  allows to formulate mathematically  precise convergence results. One can take a more pragmatic point of view that when  there is a ``small" random noise  in the observed values,   the respective test statistics    properly normalized with respect to magnitude of that noise  have approximately a noncentral chi square  distribution.
}
\end{remark}

The asymptotics of the test statistic $T_N(r)$   depends on $r$ and also on  the cardinality $m$ of the index set $\O$. Suppose now that
more observations become available  at additional entries of the matrix. That is we are testing now the model with  a larger  index set $\O'$,  of cardinality $m'$, such that   $\O\subset \O'$. In order to emphasize that the test statistic also depends on the corresponding  index set  we add the index set in the respective notations.  Note that if $Y^*$ is a solution of rank $r$ for both  sets $\O$ and  $\O'$ and the model  is globally (locally) identifiable at $Y^*$ for  the set $\O$, then the model  is   globally (locally) identifiable at $Y^*$ for  the set $\O'$. Note also that if the regularity condition \eqref{local-1} holds at $Y^*$ for the smaller model  (i.e. for $\O$), then it holds at  $Y^*$ for the larger model (i.e. for $\O'$). The following result can be proved in the same way as Theorem \ref{th-test} (cf., \cite{ste1985}).

\begin{proposition}
\label{th-test3}
Consider  index sets $\O\subset \O'$ of  cardinality
$m=|\O|$ and $m'=|\O'|$, and
  the noisy observation model  \eqref{observ}. Suppose that  the  model is globally identifiable at $Y^*\in \M_r$   and condition \eqref{local-1} holds at $Y^*$ for the smaller model (and hence for both models). Then  the  statistic $T_N(r,\O')-T_N(r,\O)$ converges in distribution to noncentral $\chi^2$ with ${\rm df}_{r,\O'}- {\rm df}_{r,\O}=m'-m$ degrees of freedom and  the  noncentrality parameter
$\delta_{r,\O'}-\delta_{r,\O}$, and $T_N(r,\O')-T_N(r,\O)$ is asymptotically independent of $T_N(r,\O)$.
\end{proposition}

{\color{black}
For   given index set $\O$  and observed  (estimated)  values $M_{ij}$, $(i,j)\in \O$, the  statistic $T_N(r)$ can be used for testing the (null) hypothesis that the ``true" rank is $r$. That is the  null  hypothesis is rejected if $T_N(r)$ is large enough on the scale of the $\chi^2$ distribution with the respective  ${\rm df}_r$  degrees of freedom. It is often observed in practice that such tests reject the null hypothesis even when the fit is reasonable. In that respect the role of values $\Delta_{ij}$ in the model is to suggest that the ``true" model is true only approximately, and the corresponding noncentrality parameter $\delta_r$ gives an indication of the deviation from  the exact rank $r$ model.
 It is a common practice to perform such tests sequentially for increasing  values of $r$,   with all deficiencies of such sequential testing.

 Such testing procedure assumes that the sample size $N$ is given and the corresponding variances $\sigma^2_{ij}$ can be consistently estimated. When the observed values are obtained by averaging $N$ data points, this is available in the straightforward way (see Remark \ref{rem-model}).
Otherwise   setting  $N=1$ and assuming that all $\sigma^2_{ij}=\sigma^2$, $(i,j)\in \O$,  are equal to each other, we need to specify range of $\sigma^2$. We will discuss this further in Section \ref{sec-numer}.
}


{\color{black}
\begin{remark}
\label{rem-asym}
{\rm
It is also possible to give asymptotic distribution of solutions of problem \eqref{weight}. Suppose now that
 the assumptions of Theorem \ref{th-test} hold with
all $\Delta_{ij}$ in equation \eqref{observ} being  zeros. Let $\hat{Y}_N$ be a solution of problem \eqref{weight}, i.e.,
\begin{equation}\label{assym-y}
\hat{Y}_N\in \arg\min_{Y\in \M_r}\sum_{(i,j)\in \O}w_{ij}\Big(\underbrace{Y^*_{ij}+\e_{ij}}_{M_{ij}}-Y_{ij}\Big)^2.
\end{equation}
Consider   operator  $\A:\bbv_\O\to \T_{\M_r}(Y^*)$ defined as
\begin{equation}\label{operator}
\A(W):=\arg \hspace{-5mm}\min\limits_{H\in \T_{\M_r}(Y^*)}\sum_{(i,j)\in \O}\sigma^{-2}_{ij}\left(W_{ij}-H_{ij}\right)^2,
\end{equation}
for $W\in \bbv_\O$.
Because of the assumption of well posedness (which is equivalent to
\eqref{sufcon})  the minimizer in \eqref{operator} is unique and hence $\A(W)$ is well defined.
Then
\begin{equation}\label{fappr}
  \hat{Y}_N= \A(M)+o_p(N^{-1/2}).
\end{equation}
Note   that the operator $\A$  is linear.

We have  that $Y^*\in \T_{\M_r}(Y^*)$ and hence $\A(P_\O(Y^*))= Y^*$.  Thus
$\A(M)= Y^* +\A(E)$, where $E\in \bbr^{n_1\times n_2}$ is such that $E_{ij}=\e_{ij}$ for $(i,j)\in \O$, and $E_{ij}=0$ otherwise.
Since   $N^{1/2}\e_{ij}$, $(i,j)\in \O$, converge  in distribution to normal with mean zero and variance $\sigma^2_{ij}$  and independent of each over, it follows that $N^{1/2}(\hat{Y}_N-Y^*)$ converges in distribution to the  random matrix $\A(Z)$, where $Z\in \bbv_\O$ is a  random matrix with entries $Z_{ij}\sim \N(0,\sigma_{ij}^2)$, $(i,j)\in \O$,  having normal distribution and independent of each over. Note that since  $\A(\cdot)$ is a linear operator, $\A(Z)$ has a multivariate normal distribution with zero means. Since $\A(Z)$ belongs to the linear subspace $ \T_{\M_r}(Y^*)$  of $\bbr^{n_1\times n_2}$, the multivariate normal distribution of $\A(Z)$ is degenerate.
}
\end{remark}

}

\section{Numerical Examples}
\setcounter{equation}{0}
\label{sec-numer}
We present some numerical experiments to illustrate our theory\footnote{More discussions can be found in a supplementary material at \textsf{https://www2.isye.gatech.edu/$\sim$yxie77/Experiment.pdf}.}. \yao{In this section, without further notification, nuclearnorm minimization is solved by TFOCS \cite{becker2011templates} in Matlab and LRMA problem is solved by 'SoftImpute' \cite{mazumder2010spectral}(regularization parameter equals to 0) in R.}
\subsection{An example of 6$\times$6 matrix considered in \cite{wilson39}}
As pointed in Remark \ref{rem-wilson}, Wilson and Worcester showed in \cite{wilson39} using analysis that there are two different locally unique solutions of rank $r^*=3$ for a $6\times 6$ matrix with the index set $\O$ corresponding to its off-diagonal elements. The matrix $M$ in that example is given by
\[M=\left(\begin{array}{cccccc}
	0 & 0.56& 0.16& 0.48& 0.24& 0.64\\
	 0.56& 0& 0.20& 0.66& 0.51& 0.86\\
	 0.16& 0.20& 0& 0.18& 0.07& 0.23\\
	 0.48& 0.66& 0.18& 0& 0.3& 0.72\\
	 0.24& 0.51& 0.07& 0.30& 0&0.41 \\
	 0.64& 0.86& 0.23& 0.72& 0.41& 0\\
\end{array}\right),
\] 
and, we aim to complete the diagonal entries of the above matrix. It can be verified that there are two rank {\bf 3} solutions by filling the diagonal entries by $(0.64, 0.85, 0.06, 0.56, 0.50,0.93)$, and 
$(0.42,0.90, 0.06, 0.55,0.39,1.00)$, respectively.

This simple test case where we know the ground truth well illustrates the problem. \yao{Both nuclear norm minimization and LRMA failed to recover any of these two local solutions above.} The soft-thresholded SVD  converges to a completely incorrect solution with off-diagonal far off from those of $M$,  
and the nuclear norm minimization produces a rank {\bf 4} solution by filling out the diagonal entries by
(0.44, 0.76, 0.05, 0.53, 0.19, 0.96).
Note that here both optimal solutions are well-posed, and yet these numerical procedures can not recover any one of them.
It is not clear how typical this example, of different locally optimal solutions, is.
Recall that generally the nuclear norm minimization problem possesses unique optimal solution. However, it is not clear how well it approximates the ``true'' minimal rank solution when it is observed with a noise.

\subsection{Probability of well-posedness}

\begin{figure}
	\includegraphics[width=0.6\linewidth]{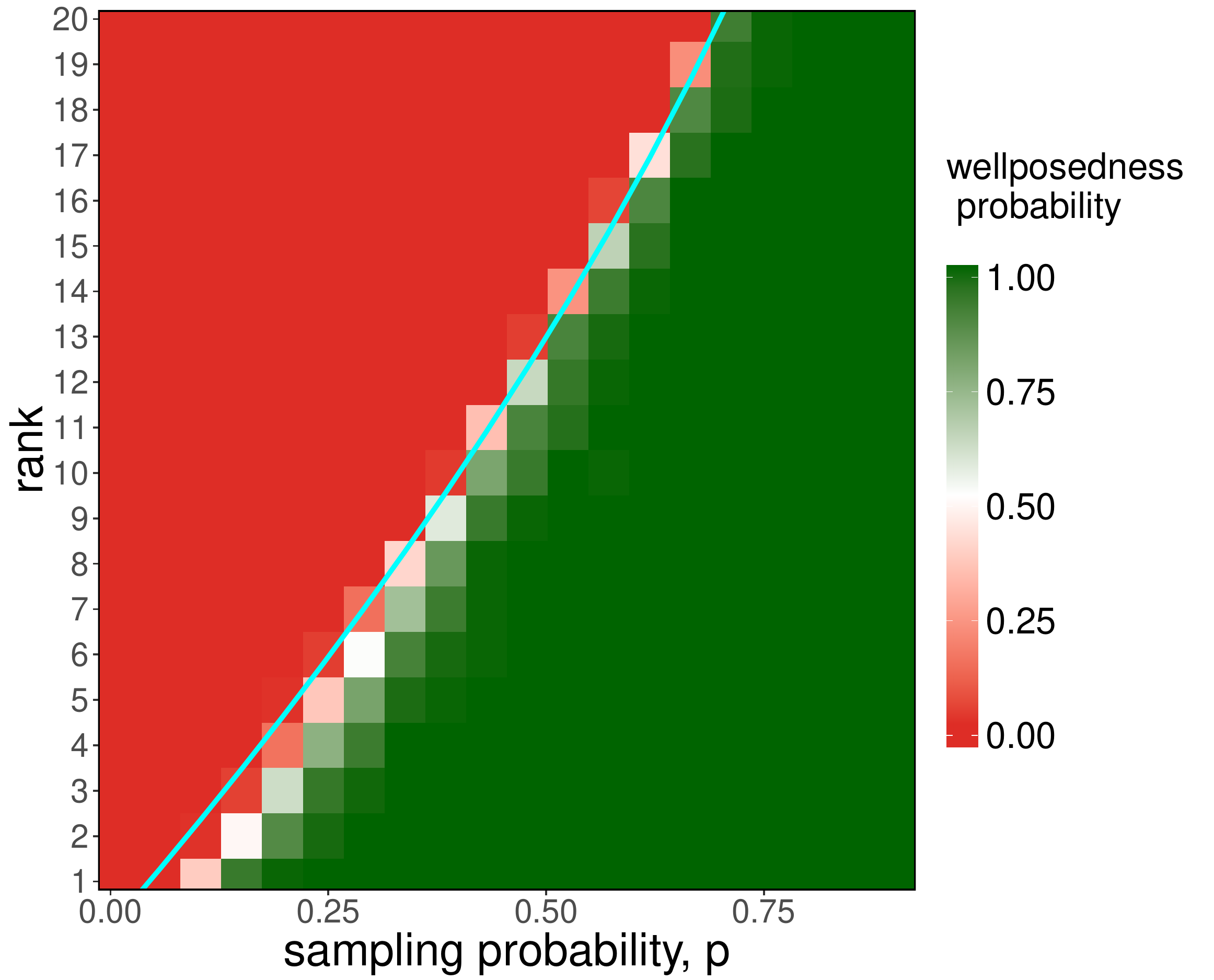}
	\centering
	\caption{Probability that well-posedness is satisfied; random instances different rank and sampling probability. For each sampling probability and rank, we generate $Y^*$ and $\O$. Then, we check the well-posedness condition and compute the probability. Blue curve is the estimated generic bound for the corresponding sampling probability.}
	\label{wlpProb}
\end{figure}

\yao{We show the probability of satisfying the well-posedness condition, for generating random cases. For each rank $r^*$, we generate an $40\times r^*$ orthonormal matrix $V$, an $50\times r^*$ orthonormal matrix $W$, and an $r^*\times r^*$ diagonal matrix $D$ and setting $Y^*=VDW^\top$. For each instance, we randomly generate the observation pattern $\Omega$ such that each entry is observed with probability $p$. We check the well-posedness condition according to Theorem \ref{pr-uniqa} and using the verifiable algebraic condition. Repeat this 100 times and compute the percentage of cases that satisfy the well-posedness condition. Figure \ref{wlpProb} shows the resulted proportion. We also plotted the generic bound, the estimation $\hat\cR(n_1,n_2,p)=(n_1+n_2)/2-((n_1+n_2)^2/4- n_1 n_2p)^{1/2}$. Figure \ref{wlpProb} shows that the probability for a matrix to satisfy the well-posedness condition is not small when the true rank is less than the estimated generic lower bound and the probability converge to 1 fast when the rank is 2 or 3 less than the generic bound. This demonstrates that the $\hat\cR(n_1,n_2,p)$ is a sharp bound. }

\subsection{Comparison of LRMA and nuclear norm minimization}

In this section, we compare the performance of LRMA and matrix complemtion using standard nuclear norm minimization, when well-posedness condition is satisfied and when it is violated, respectively. The results show that the well-posedness condition is indeed necessary for good recovery performance. Moreover, our examples show that LRMA performs more stable than nuclear norm minimization in these cases. 

We  generate $Y^*$, an $n_1\times n_2$ matrix of rank $r^*$, by uniformly generated an $n_1\times r^*$ matrix $V$, an $n_2\times r^*$ matrix $W$ and an $r^*\times r^*$ diagonal matrix $D$ and setting $Y^* = \tilde VD\tilde W^\top$, where $\tilde V$ and $\tilde W$ are orthonormalization of $V$, $W$, respectively. We again sample $\O$ uniformly random with probability $p$, where $|\O| = m$. Observation matrix $M$ is generated by $M_{ij} = Y^*_{ij} + \varepsilon_{ij}, (i,j)\in\O$, where $\varepsilon_{ij}\sim N(0, \sigma^2N^{-1})$. Algorithms stop when either relative change in the Frobenius norm between two successive estimates,  $\|Y^{(t + 1)} - Y^{t}\|_F/\|Y^{(t)}\|_F$, is less than some tolerance, denoted as $tol$ or the number of iterations exceeds the maximum $it$. 

\subsubsection{Elementwise error for three cases}
\begin{figure}[h!]
	\includegraphics[width = \linewidth]{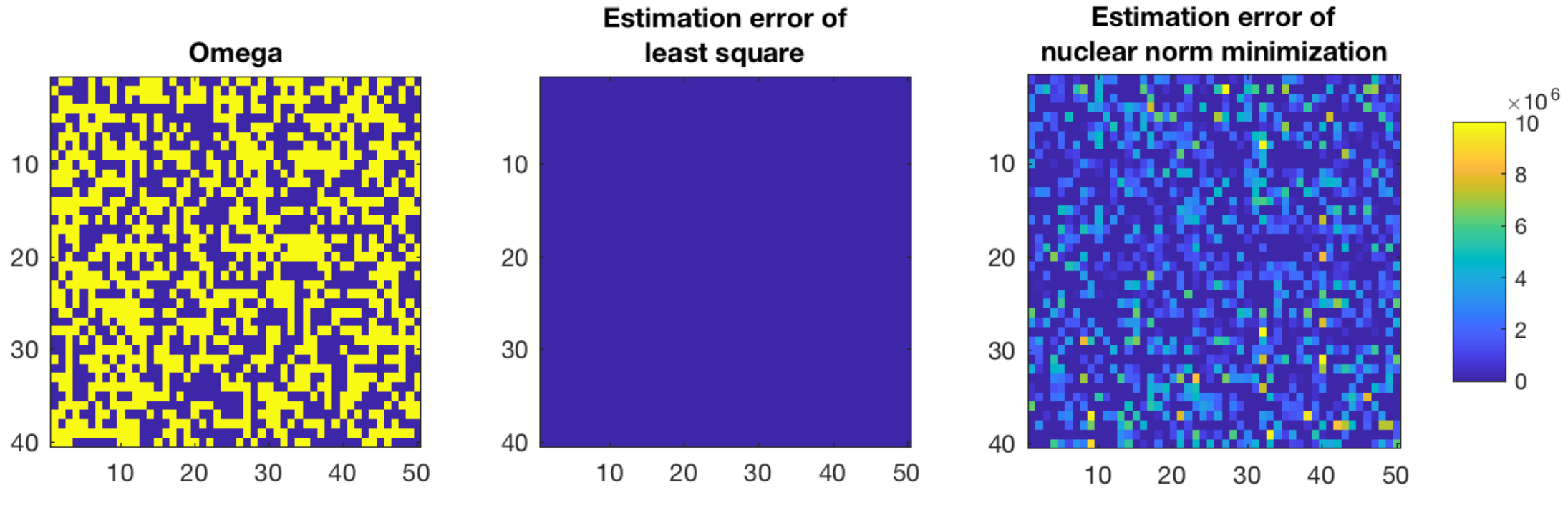}
	\centering
	\caption{When well-posedness is satisfied, absolute errors at each entries $|Y_{ij} - Y_{ij}^*|$  for the LRMA (middle panel) and the nuclear norm minimization (right panel) methods. The left panel show the sampling pattern $\Omega$. 
	 Here the true matrix $Y^* \in \mathbb{R}^{40\times50}$, $\rank(Y^*) = 10$, $|\O| = 1000$, $\varepsilon_{ij}\sim N(0, 5^2/50)$ and the observation matrix $M_{ij} = Y^*_{ij} + \varepsilon_{ij}, (i,j)\in\O$.}
	\label{Con_bl}
\end{figure}

We first consider three individual instances, when the well-posedness condition is satisfied and violated, respectively: \\
(1) In Figure \ref{Con_bl} the well-posedness condition is satisfied. The element-wise reconstruction error for LRMA is much smaller than that of the nuclear norm minimization. In this experiment, $n_1 = 40$, $n_2 = 50$, $r^* = 10$, $m = 1000$, $\sigma = 5$, $N = 50$ and $\O$ is sampled until well-posedness condition is satisfied. The parameters are $tol = 10^{-20}$ and $it = 50000$. 


\begin{figure}[h!]
	\includegraphics[width = \linewidth]{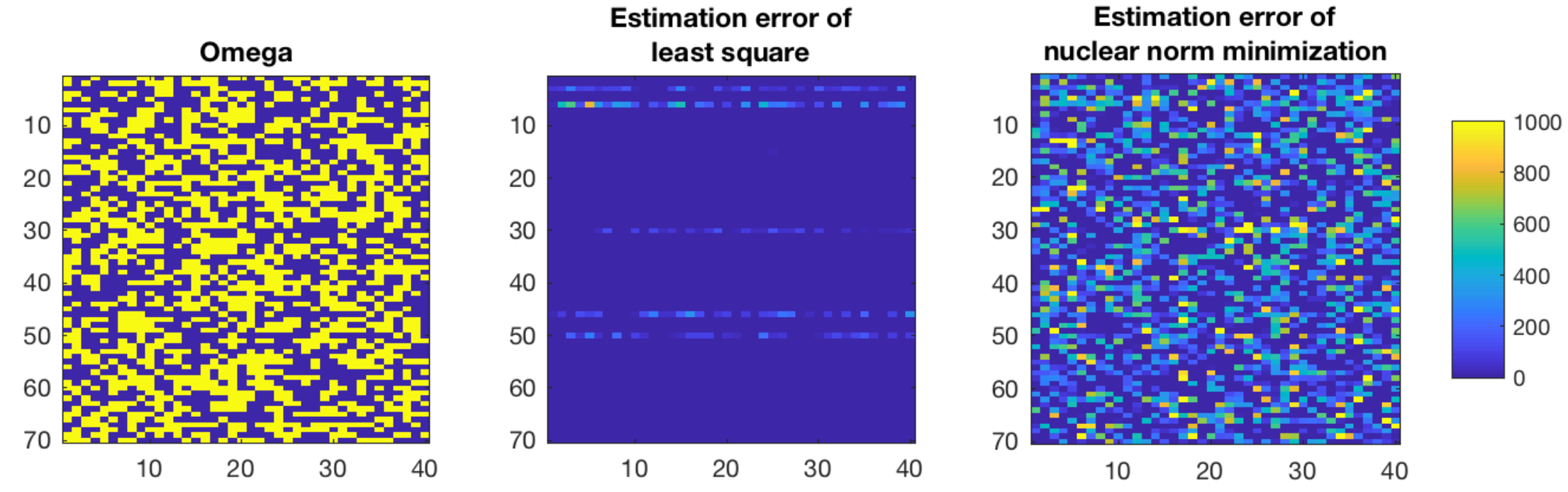}
	\centering
	\caption{When well-posedness is violated, absolute errors at each entries $|Y_{ij} - Y_{ij}^*|$  for the LRMA (middle panel) and the nuclear norm minimization (right panel) methods. The left panel show the sampling pattern $\Omega$. 
	 Here the true matrix $Y^* \in \mathbb{R}^{70\times40}$, $\rank(Y^*) = 11$, $|\O| = 1300$, $\varepsilon\sim N(0, 5^2/50)$ and the observation matrix $M_{ij} = Y^*_{ij} + \varepsilon_{ij}, (i,j)\in\O$. The necessary condition for well-posedness is violated (i.e. the numbers of observations are less than 11),  at row with index numbers 3, 6, 30, 46, 50.
	}
	\label{Con_rcon}
\end{figure}

\noindent (2) In Figure \ref{Con_rcon}, the well-posedness condition is violated.  As predicted by our theory, both LRMA and nuclear performs worse, and the errors are especially large at index numbers 3, 6, 30, 46, 50, where the necessary condition for well-posedness is violated. Still, in this situation, nuclear norm minimization has larger total recover error than LRMA. In this experiment, $n_1 = 70$, $n_2 = 40$, $r^* = 11$, $m = 1300$, $\sigma = 5$, $N = 50$. We repeatedly sample $\O$ until the necessary condition for well-posedness is violated to generate our instances. The parameters $tol = 10^{-16}$ and $it = 50000$.  

\begin{figure}[h!]
	\includegraphics[width = \linewidth]{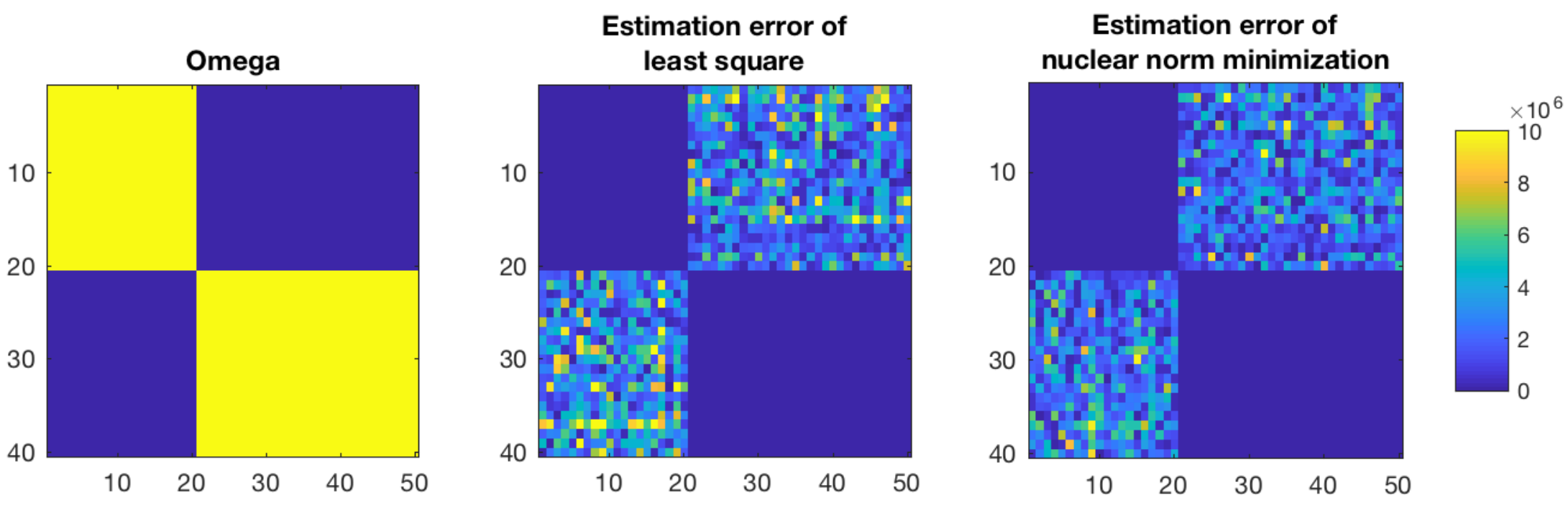}
	\centering
	\caption{When $\O$ is reducible, absolute errors at each entries $|Y_{ij} - Y_{ij}^*|$  for the LRMA (middle panel) and the nuclear norm minimization (right panel) methods. The left panel show the sampling pattern $\Omega$. 
	 Here true matrix $Y^* \in \mathbb{R}^{40\times50}$, $\rank(Y^*) = 10$, $|\O| = 1000$, $\varepsilon_{ij} \sim N(0, \frac{5^2}{50})$ and the observation matrix $M_{ij} = Y^*_{ij} + \varepsilon_{ij}, (i,j)\in\O$. $\Omega$ is reducible. In this case, only two diagonal block matrices $M_1 \in \mathbb{R}^{20\times20}$ and $M_2 \in\mathbb{R}^{20\times30}$ are observed.
	}
	\label{Con_red}
\end{figure}

\noindent (3) In Figure \ref{Con_red}, $\O$ is reducible and thus the well-posedness condition is violated. Consistent with our theory, in this situation, both methods fail to recover the true matrix since the necessary condition of local uniqueness is violated. 
In this experiment, $n_1 = 40$, $n_2 = 50$, $r^* = 10$, $m = 1000$, $\sigma = 5$, $N = 50$ and $\O = \{(i,j)\in\{1\cdots20\}\times\{1\cdots20\} \cup \{21\cdots 40\}\times\{21\cdots 50\}\}$. The parameters are $tol = 10^{-20}$ and $it = 50000$. 

\subsubsection{Mean-square-error performance}
\begin{figure}[h!]
\includegraphics[width = 0.6\linewidth]{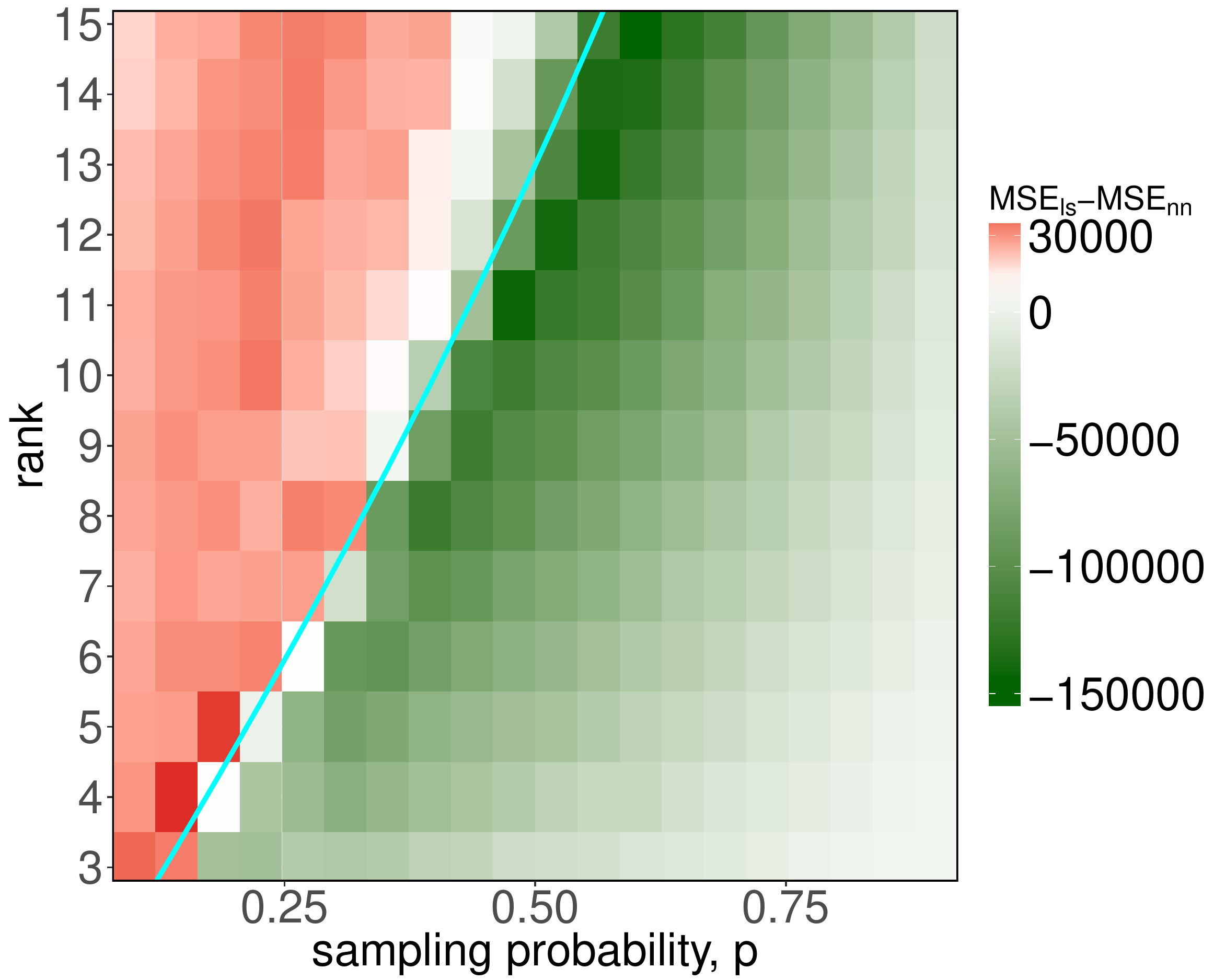}
\centering
\caption{Difference between MSE of LRMA and nuclear norm minimization. The blue curve is the estimated generic bound for the corresponding sampling probability.}
\label{MSE_ls_nn}
	
\end{figure}

In this section, we consider the mean-square-error performance, defined by 
\[{\rm MSE} = \frac{1}{n_1n_2 K} \sum_{k=1}^K \sum_{i,j} (Y_{ij,k}^* - \hat{Y}_{ij,k})^2 \]
where $K$ is the total number of repetitions. 
\yao{Figure \ref{MSE_ls_nn} shows the difference between the \textit{mean square error} of LRMA and nuclear norm minimization. In this experiment, $n_1 = 40$, $n_2=50$, $\sigma=5$, and we generate 50 random instances to compute the average error. 
The estimated $\hat\cR(n_1, n_2,p)$ is also drawn as the blue curve. Figure \ref{MSE_ls_nn} shows that, indeed, as predicted by our theory, when the true rank is lower than the estimated generic lower bound, the performance of LRMA is much better than nuclear norm minimization.
}

\subsection{Testing for true rank}
\subsubsection{Asymptotic distribution of test statistic}

In Section \ref{sec-stat} (see \eqref{observ}), we show that the asymptotical distribution of the test statistic for the ``true'' rank is $\chi^2$ distribution, which we will verify numerically here.  
We generate the true matrix $Y^*$, an $n_1\times n_2$ matrix of rank $r^*$, by uniformly generated an $n_1\times r^*$ matrix $V$, an $n_2\times r^*$ matrix $W$, and an $r^*\times r^*$ diagonal matrix $D$ and setting $Y^* = \tilde VD\tilde W^\top$, where $\tilde V$ and $\tilde W$ are orthonormalization of $V$, $W$, respectively. We sample $\O$ uniformly random, where $|\O| = m$. The noisy and repeated observation matrices are generated by $M^{(k)}_{ij} = Y^*_{ij} + \varepsilon^{(k)}_{ij}, (i,j)\in\O$, where $\varepsilon^{(k)}_{ij}\sim N(0, \sigma^2N^{-1})$. In computing the test statistic  $T_N^{(k)}(r)$ \eqref{stat-1}, the least square approximation is solved by a soft-threshholded SVD solver. The algorithm stops when either relative change in the Frobenius norm between two successive estimates, is less than some tolerance, denoted as $tol$ or the number of iterations reaches the maximum, denoted as $it$.

\begin{figure}[h!]
	\includegraphics[width = 0.6\linewidth]{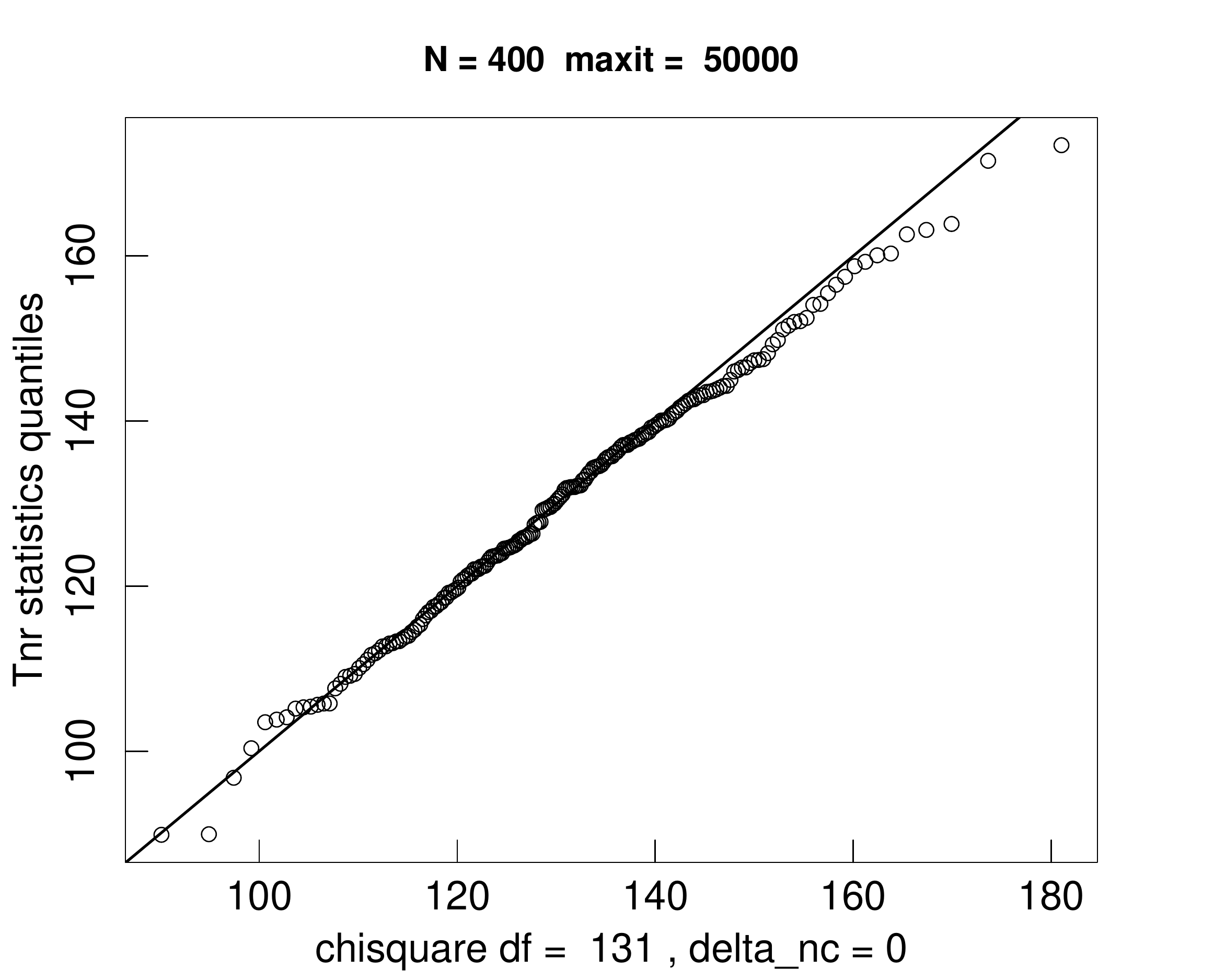}
	\centering
	\caption{Q-Q plot of $T_N(r)$ against quantiles of $\chi^2$ distribution: $Y^*\in\mathbb{R}^{40\times50}$, $rank(Y^*) = 11$, $|\O| = 1000$, the observation matrix $M$ is generated 200 times, $M^{(k)}_{ij} = Y^*_{ij} + \varepsilon^{(k)}_{ij}, (i,j)\in\O$, where $\varepsilon^{(k)}_{ij}\sim N(0,\frac{5^2}{400})$. For each $M^{(k)}$, $T^{(k)}_N(r)$ is computed as equation \ref{stat-1}. By Theorem \ref{th-test}, $\{T^{(k)}_N(r)\}$ follows central $\chi^2$ distribution with the degree-of-freedom ${\rm df}_r =m-r(n_1+n_2-r) = 131$.}
	\label{chisq}
\end{figure}

Figure \ref{chisq} shows the Q-Q plot of $\{T_N^{(k)}(r)\}^{200}_{k=1}$ against the corresponding $\chi^2$ distribution. In this experiment, $n_1 = 40$, $n_2 = 50$, $r^* = 11$, $m = 1000$, $\sigma = 5$, $N = 400$ and $\O$ is sampled until well-posedness condition is satisfied. The parameters $tol = 10^{-20}$ and $it = 50000$. From the result, we can see $T_N(r)$ follows the central $\chi^2$ distribution with  a degree of freedom ${\rm df}_r =m-r(n_1+n_2-r) = 131$, which is consistent with Theorem \ref{th-test}.

\begin{figure}
	\includegraphics[width = 0.6\linewidth]{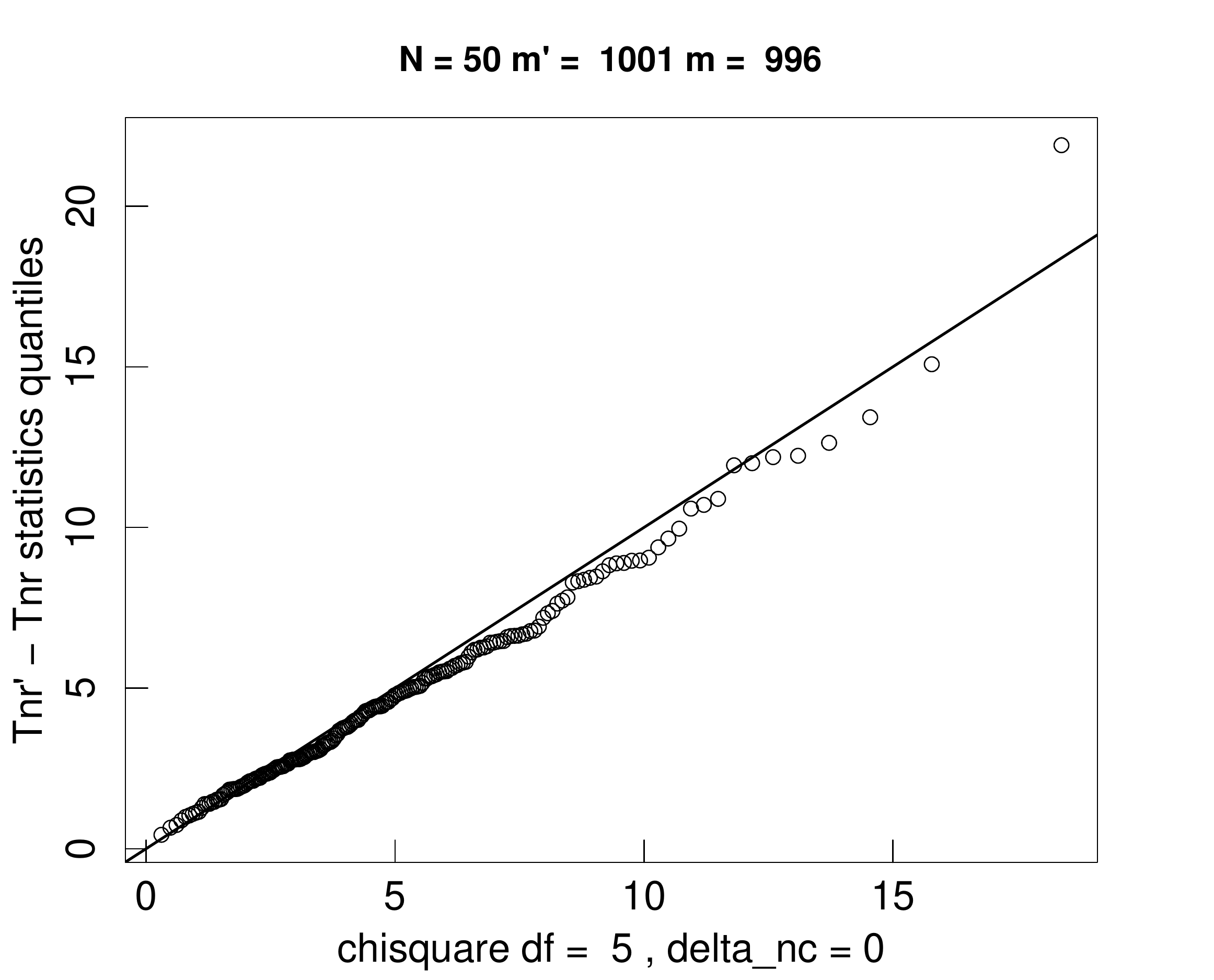}
	\centering
	\caption{Q-Q plot of $T_N(r,\O') - T_N(r,\O)$ against the quantiles of $\chi^2$ distribution:  $Y^*\in\mathbb{R}^{40\times50}$, $rank(Y^*) = 11$, $|\O'| = 1001$, $|\O|=996$, where $\O\subset\O'$. The observation matrix $M'$ and $M$ are generated 200 times,  
	By Theorem \ref{th-test3}, $\{T^{(k)}_N(r, \O') - T^{(k)}_N(r,\O)\}$ follows central $\chi^2$ distribution with the degree-of-freedom ${\rm df}_{r,\O'}- {\rm df}_{r,\O}=m'-m = 5$ .}
	\label{diffO}
\end{figure}

Figure \ref{diffO} shows the Q-Q plot of $\{T_N^{(k)}(r, \O') - T_N^{(k)}(r,\O)\}^{200}_{k=1}$ against the corresponding $\chi^2$ distribution. In this experiment, $n_1 = 40$, $n_2 = 50$, $r^* = 11$, $m = 996$, $\sigma = 5$, $N = 50$, $m' = |\O'| = 1001$ and $\O$ is sampled until well-posedness condition is satisfied (note that $\O'$ also satisfied well-posedness condition since $\O'^C \subset\O^C$). The parameters $tol = 10^{-20}$ and $it = 50000$. From the result, we can see $T_N(r, \O') - T_N(r,\O)$ follows a central $\chi^2$ distribution with a degree of freedom ${\rm df}_{r,\O'}- {\rm df}_{r,\O}=m'-m = 5$, which is consistent with Theorem \ref{th-test3}.



\subsubsection{Test for true rank}
\yao{
As discussed in Section \ref{sec-stat}, we can  determine the true rank $r^*$ by sequential $\chi^2$ tests. That is, for $r$ ranging from $1$ to $\left\lceil \mathfrak{R}(n_1, n_2, m)\right\rceil$, we solve the least square approximations and compute $T_N(r)$. According to $T_N(r)$ we can determine which rank can be accepted for a predefined significant level. Table \ref{table:sim} shows a result of sequential rank test on a simulated data set. In this experiment, $n_1 = 40$, $n_2 = 50$, $r^* = 9$, $m = 1000$, $\sigma = 5$, $N = 100$, and $\O$ is sampled until well-posedness condition is satisfied. The true rank 9, is the first one accepted for 0.05 significant level.}
\begin{table}
	\caption{$p$-value for sequential rank test in simulation.}
\centering
{\small
\begin{tabular}{|c|c||c|c|}\hline
	rank & p-value  &rank & p-value \\\hline
	1& 0.00&  7& 0.00\\\hline
	2& 0.00&  8& 0.00\\\hline
	3& 0.00&  \bf 9& \bf 0.94 \\\hline
	4& 0.00&  10& 0.69\\\hline
	5& 0.00&  11&0.41\\\hline
	6& 0.00&  12&0.00\\\hline	
\end{tabular}}
\label{table:sim}
\end{table}

\begin{figure}[h!]
\includegraphics[width = .6\linewidth]{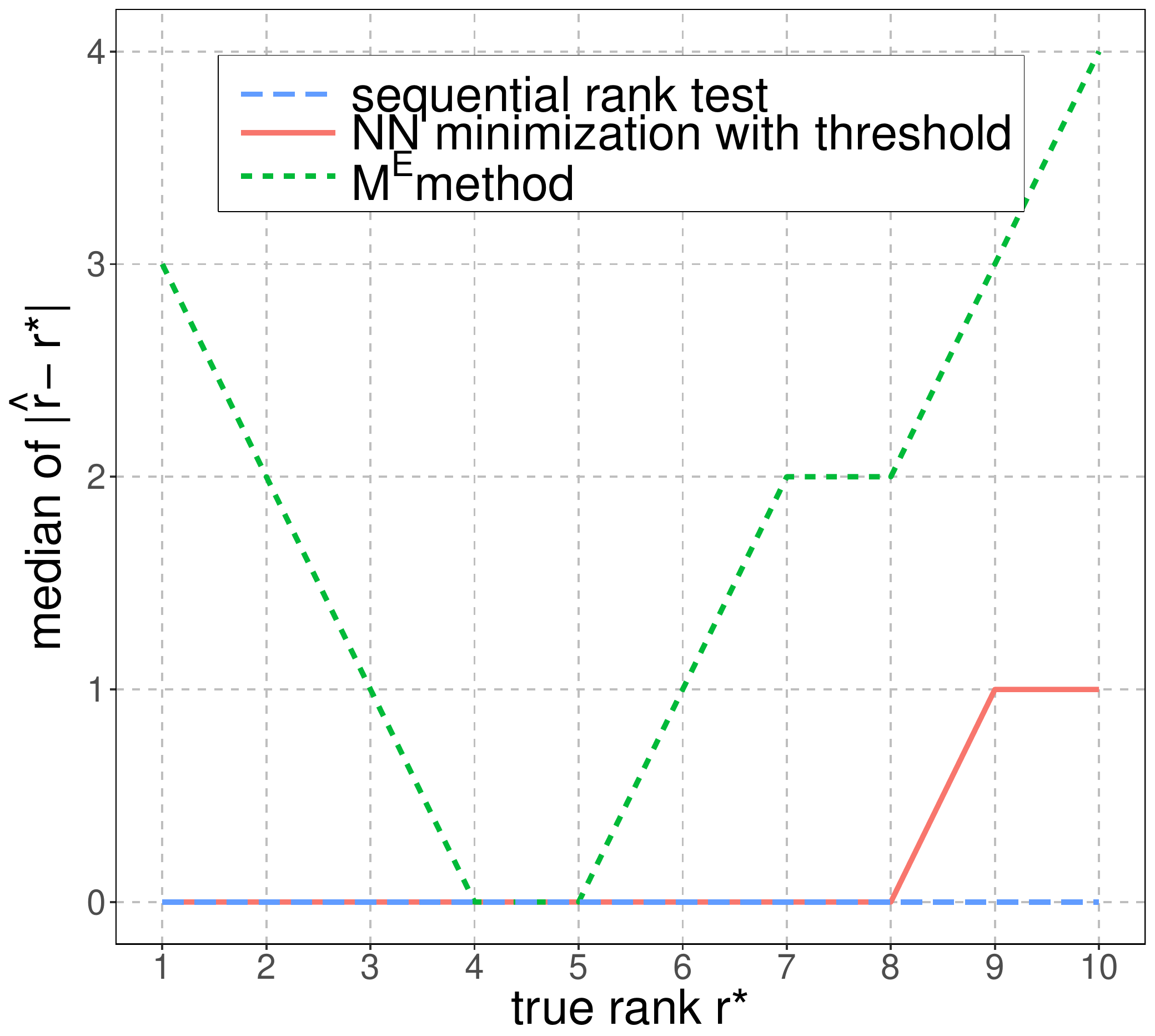}
\centering
\caption{Comparison of rank selection between sequential $\chi^2$ test, nuclear norm minimization and $M^E$ method, sampling probability $p$=0.3. For each true rank, we compute the median of rank error for 100 experiments. $Y^{*(k)}\in\mathbb{R}^{100\times 1000}$, $M^{(k)}_{ij} = Y^{*(k)}_{ij} + \varepsilon^{(k)}_{ij}, (i,j)\in\O$, where $\varepsilon^{(k)}_{ij}\sim N(0,\frac{5^2}{50})$. Threshold $b_{nm} = 0.25$, $b_{ME} = 0.13$ for nuclear norm minimization and $M^E$ method, respectively.}
\label{rankSeleComp}
	
\end{figure}

Figure \ref{rankSeleComp} shows the comparison of rank selection between our sequential rank test, nuclear norm minimization and the method suggested in \cite{keshavan2010matrix} (we refer it as $M^E$ method in the following). Since the nuclear norm minimization and $M^E$ method can't give us the exact rank, we choose the rank by thresholding the percentage of the singular value of the recovered matrix in this two methods, i.e. $\hat r = \argmin_r\sum_{i=1}^r \lambda_{(i)}/\sum^{min(n_1,n_2)}_{i=1}\lambda_{(i)} > b$, where b is some threshold. \yao{In this experiment, $n_1 =100$, $n_2=1000$, $\sigma = 5$, $N=50$ and the sampling probability $p=0.3$.} For each true rank, we generate 100 instances of $(Y^*, \O, M)$, complete the rank selection with these three methods and compute the median of the error of estimated rank of each method. \yao{For the sequential rank test, we choose the first rank accepted with 0.05 significant level.} For nuclear norm minimization and $M^E$ method, we choose the threshold that gives us the best results for these two methods. It shows that selection by sequential $\chi^2$ test outperforms the other two methods.


\section{Conclusion}\label{sec:conclusion}

In this paper, we have examined the matrix completion from a geometric viewpoint and established a sufficient  condition for local uniqueness of solutions. Our characterization assumes deterministic patterns and the results are general. We argue that the exact  minimum rank matrix completion (MRMC) leads to  either
 unstable or non-unique solutions and thus the alternative low-rank matrix approximation (LRMA) is a more reasonable  approach. We propose  a  statistical test for rank selection, based on observed entries, which can be useful for  practical matrix completion algorithms. Assuming the model \eqref{observ}, it is also possible to derive asymptotic of the optimal value and, under rather stringent conditions, of the optimal solutions  of the minimum trace (MT)  problem \eqref{sim-2} (cf., \cite{AShapiro_2017}).

\yao{For small values of the ``true" rank, when the respective  dual of the ``true" MT   problem has more than one optimal solution,   the asymptotic bias of the optimal value of the approximating MT problem  is of  order $O(N^{-1/2})$  (see Remark \ref{rem-as2}).
 On the other hand, under the model  \eqref{observ} when the values $M_{ij}$, $(i,j)\in \O$,  are computed by averaging $N$ data points having normal distribution (see Remark \ref{rem-model}), the least squares approach corresponds to the Maximum Likelihood method which is an asymptotically efficient estimation procedure.
 This gives an insight into  the relatively  poor performance of the nuclear norm approach, as compared with the least squares method, as reported in Section \ref{sec-numer}.
}

\bibliographystyle{ieee}

\bibliography{references,Poisson_MC,mc}


\appendix

{\bf Proof of Theorem \ref{pr-uniq}}
We argue by a contradiction. Suppose that there is a sequence $\{Y_k\}\subset \M_r$   (with $Y_k\ne\bar{Y}$) converging to  $\bar{Y}$ such that $P_\O(Y_k)=M$. It follows that $Y_k-\bar{Y}\in  \bbv_{\O^c}$. By passing to a subsequence if necessary we can assume that $(Y_k-\bar{Y})/t_k$,
 where $t_k:=\|Y_k-\bar{Y}\|$,
 converges to some $H\in \bbv_{\O^c}$. Note that  $H\ne 0$. Moreover
 $Y_k=\bar{Y}+t_k H+o(t_k)$,  and hence $H\in \T_{\M_r}(\bar{Y})$.
{\color{black} That is  $H\in \bbv_{\O^c}\cap \T_{\M_r}(\bar{Y})$,   and   $H\ne 0$ by the construction. This gives the desired contradiction with \eqref{local-1}.}
 \vspace{.1in}\\
{\color{black}
{\bf Proof of Theorem \ref{th-wpgen}}
Let $\varrho$ be the  characteristic rank of
mapping $\cF$.  Consider $\theta^*\in \Theta$ such that  $\varrho=\rank\big (\Delta(\theta^*)\big)$. It follows that matrix $\Delta(\theta^*)$ has an $\varrho\times \varrho$ submatrix whose determinant is not zero. Consider function $\phi:\Theta\to \bbr$ defined as the determinant  of the corresponding
$\varrho\times \varrho$ submatrix of $\Delta(\theta)$. We have that $\phi(\cdot)$ is a polynomial function and is not identically zero on $\Theta$ since by the construction $\phi(\theta^*)\ne 0$. Since   $\Theta$ is connected, it follows that the set $\{\theta\in \Theta:\phi(\theta)=0\}$ is ``thin", in particular has Lebesgue measure zero. That is, $\phi(\theta)\ne 0$
and hence $\rank\big (\Delta(\theta)\big)\ge \varrho$
 for a.e. $\theta\in \Theta$. Also by the definition of $\varrho$ we have that  $\rank\big (\Delta(\theta)\big)\le \varrho$  for all $\theta\in \Theta$.  It follows that $\rank\big (\Delta(\theta)\big)= \varrho$  for a.e. $\theta\in \Theta$. Since rank of $\Delta(V,W,X)$ is the same for all  $X\in \bbv_{\O^c}$, this completes the   proof of the  assertion (i).
Since $\rank\big (\Delta(\cdot)\big)$ is a lower semicontinuous function, the  assertion (ii) follows.

Now consider a regular point $\bar{\theta}=(\bar{V},\bar{W},\bar{X})$  with $\bar{X}=0$,  and  the corresponding matrix  $\bar{Y}=\bar{V}\bar{W}^\top$.
Since $\bar{\theta}$ is regular, we have that  rank of $\Delta(\theta)$ is constant (equal $\varrho$) for all $\theta$ in a neighborhood of $\bar{\theta}$.
By the Constant Rank Theorem it follows that there is a neighborhood $\V$ of $\bar{\theta}$ such that the set  $\s:=\{\cF(\theta):\theta\in \V\}$ forms a smooth manifold of dimension $\varrho$ in $\bbr^{n_1\times n_2}$.
The tangent space to this manifold at $\bar{Y}$ is the space
$ \T_{\M_r}(\bar{Y})+\bbv_{\O^c}$.
Hence if $\varrho=\cf(r,m)$, then
\[
\dim\left( \T_{\M_r}(\bar{Y})+\bbv_{\O^c}\right)= \dim( \T_{\M_r}(\bar{Y}))+\dim (\bbv_{\O^c}).
\]
Consequently
$\dim\left(\T_{\M_r}(\bar{Y})\cap \bbv_{\O^c}\right)=0$, and thus
condition \eqref{local-1} follows. On the other hand if $\varrho<\cf(r,m)$, then the manifold
$ (\bbv_{\O^c}+\bar{Y})\cap \M_r$,  in a neighborhood of  $\bar{Y}$,  has a positive dimension.  Thus  in that case the solution of MRMC is not locally unique and
condition \eqref{local-1} does not hold. This completes the  proof of the assertions (iii) and (iv).
  }
  \vspace{.1in}\\
{\bf Proof of Theorem \ref{pr-reduc}}
Suppose that $\O$ is   reducible. Then by making  permutations of rows and columns if necessary,  it can be assumed that $M$ has the   block diagonal form as in (\ref{M_block}).
Let $\bar{Y}$ be a respective minimum rank solution. That is
$M_1=V_1 W_1^\top$, $M_2=V_2 W_2^\top$ and
$\bar{Y}=V W^\top$ with $V=\begin{psmallmatrix}
  V_1 \\
 V_2
  \end{psmallmatrix}
$ and
$W=\begin{psmallmatrix}
 W_1 \\
W_2
\end{psmallmatrix}$
being $n_1\times r$ and $n_2\times r$  matrices of rank $r$. 
Note that
$\bar{Y}=\begin{psmallmatrix}
  M_1 &   V_1 W_2^\top \\
V_2 W_1^\top & M_2
\end{psmallmatrix}.$
By changing $V_1$ to $\alpha V_1$ and $W_1$ to $\alpha^{-1} W_1$ for $\alpha\ne 0$, we change matrix $\bar{Y}$ to matrix
$\begin{psmallmatrix}
  M_1 &   \alpha V_1 W_2^\top \\
\alpha^{-1}V_2 W_1^\top & M_2\end{psmallmatrix}
$.
If  $V_1 W_2^\top\ne 0$ or   $V_2 W_1^\top\ne 0$, we obtain that
 solution $\bar{Y}$ is not locally unique. {\color{black} On the other hand when  both $V_1 W_2^\top= 0$ and    $V_2 W_1^\top= 0$, and hence
 $\bar{Y}=\begin{psmallmatrix}
  M_1 &   0 \\
0 & M_2
\end{psmallmatrix}$,
rank $r$ solutions for example  are matrices of the form $\bar{Y}=\begin{psmallmatrix}
  M_1 &   M_3 \\
0 & M_2
\end{psmallmatrix}$, where columns of matrix $M_3$ are linear combinations of columns of matrix $M_1$. If $M_1=0$, then we can use matrix
$\bar{Y}=\begin{psmallmatrix}
  M_1 &   0 \\
M_3 & M_2
\end{psmallmatrix}$ in the similar way.
Hence
nonuniqueness of rank $r$ solutions  follows. }
\vspace{.1in}\\
{\bf Proof of Theorem \ref{pr-rankone}}
Suppose  that $\O$ is irreducible. Consider a rank one  solution $\bar{Y}=v w^\top$ with respective vectors $v=(v_1,...,v_{n_1})^\top$ and $w=(w_1,...,w_{n_2})^\top$.
 We can assume that $v_1$ is fixed, say    $v_1=1$.
 Consider an element $M_{1 j_1}$, $(1,j_1)\in \O$, in the first row of matrix $M$.  {\color{black} Since it is assumed that each row has at least one observed entry, such element exists. }
 Since $M_{1 j_1}=v_1 w_{j_1}$, it follows that the component $w_{j_1}$ of vector $w$ is uniquely defined. Next consider element $M_{i_1,j_1}$, $(i_1,j_1)\in \O$. Since $M_{i_1 j_1}=v_{i_1} w_{j_1}$, it follows that the component $v_{i_1}$ of vector $v$ is uniquely defined.
 We proceed now iteratively. Let $\nu\subset \{1,...,n_1\}$ and $\w\subset \{1,...,n_2\}$ be  index sets for which the respective components of vectors $v$ and $w$ are already uniquely defined. Let  $j \not \in \w$ be  such that  there is $(i,j')\in\O$ with    $j'\in \w$  and hence $w_{j'}$ is already uniquely defined. Since  $M_{ij}=v_i w_j$ and $M_{ij'}=v_i w_{j'}$, it follows that   $ w_{j}$ is uniquely defined and $j$ can be added to the index set $\w$.
 If such column $j$ does not exist, take row  $i\not \in \nu$ such that there is $(i',j)\in \O$ with $i'\in \nu$. Then $v_i$ is uniquely defined   and hence   $i$ can be added  to $\nu$.
 Since $\O$ is irreducible,
  this process can be continued until all components of vectors $v$ and $w$ are uniquely defined.
\vspace{.1in}\\
{\bf Proof of Proposition \ref{pr-opt}}
Consider function defined in (\ref{func-f}).
The differential of $f(Y)$ can be written as
\[
{\rm d} f(Y)=\tr[(P_\O(Y)-M)^\top {\rm d}Y].
 \]
 Therefore  if $Y\in \M_r$ is an optimal solution of the least squares problem \eqref{leastsq}, then $\nabla f(Y)=P_\O(Y)-M$ is orthogonal to the tangent space $T_{\M_r}(Y)$.
By \eqref{orthog-2} this implies optimality conditions \eqref{optnes-1} .
\vspace{.1in}\\
{\bf Proof of Proposition \ref{pr-locuniq}}
Consider function $\phi$ defined in \eqref{func-g}, and  the problem of minimization of   $\phi(Y,\Theta)$ subject to $Y\in \M_r$ with $\Theta$ viewed as a parameter. Locally for $Y$ near $\bar{Y}\in \M_r$ the manifold $\M_r$ can be represented by a system of $K=n_1n_2-\dim(\M_r)$
equations  $g_i(Y)=0$, $i=1,...,K,$  for an appropriate smooth mapping $g=(g_1,...,g_K)$.
That is, the above optimization problem can be written as
\begin{equation}\label{problocal}
 \min  \phi(y,\theta)\;\;{\rm subject\;to}\;g_i(y)=0,\;i=1,...,K,
\end{equation}
where with some abuse of the notation we write this in terms of vectors $y=\vect(Y)$ and $\theta=\vect(\Theta)$.
Note that the mapping $g$ is such that the  gradient vectors $\nabla g_1(\bar{y}),...,\nabla g_K(\bar{y})$ are linearly independent.

 First order optimality conditions for problem \eqref{problocal} are
\begin{equation}\label{prob-2}
\nabla_y L(y,\lambda,\theta)=0,\;g(y)=0,
\end{equation}
where $L(y,\lambda,\theta):= f(y,\theta)+\lambda^\top g(y)$ is the corresponding Lagrangian.  For $\theta=\theta_0$ this system has   solution $\bar{y}$ and the corresponding vector $\bar{\lambda}=0$ of Lagrange multipliers.
We can view \eqref{prob-2} as a system of (nonlinear) equations in $z=(y,\lambda)$ variables.

We would like now to apply the Implicit Function Theorem  to this system of equations to conclude that for all $\theta$ near $\theta_0$ it has unique solution near $\bar{z}=(\bar{y},\bar{\lambda})$. Consider  the Jacobian matrix
$\begin{psmallmatrix}
  H &  G \\
G^\top & 0\end{psmallmatrix}
$
of the system \eqref{prob-2} at $(y,\lambda)=(\bar{y},\bar{\lambda})$,
 where $H:=\nabla_{yy} \phi(\bar{y},\theta_0)$ is the Hessian matrix of the objective function  and $G:=\nabla g(\bar{y})=\left[\nabla g_1(\bar{y}),...,\nabla g_K(\bar{y})\right]$. We need to verify that this
 Jacobian matrix is nonsingular.
 This is implied by condition  \eqref{local-1}, which is equivalent to condition \eqref{sufcon}. Indeed suppose that
 \begin{equation}\label{jacob}
 \left[
  \begin{array}{ccc}
  H &  G \\
G^\top & 0
  \end{array}
  \right]
  \left[
  \begin{array}{ccc}
 v \\
u
  \end{array}
  \right]=0,
 \end{equation}
 for some vectors $v$ and $u$ of appropriate dimensions.
This means  that $Hv+Gu=0$ and  $G^\top v=0$. It follows that
  $v^\top H v=0$.
Condition $G^\top v=0$ means that $v$ is orthogonal to the tangent space $\T_{\M_r}(\bar{y})$. It follows then by  condition \eqref{sufcon}  that $v=0$. Then $Gu=0$ and hence, since $G$ has full column rank, it follows that  $u=0$.
 Since equations  \eqref{jacob} have only zero solution, it follows that this Jacobian matrix is nonsingular.
 Now by implying the Implicit Function Theorem to the system \eqref{prob-2} we obtain the required result.
 This completes the proof.
\vspace{.1in}\\
{\bf Proof of Proposition \ref{th-test}}
Note that under the specified assumptions, $M_{ij}-Y^*_{ij}$ are of stochastic order $O_p(N^{-1/2})$.
We have by Proposition \ref{pr-cons} that   an optimal solution  of problem \eqref{weight} converges in probability to $Y^*$. By the standard theory of least squares (e.g., \cite[Lemma 2.2]{sha-bio}) we can write the following local  approximation near $Y^*$ as \eqref{stat-2}.
It follows that the limiting distribution of $T_N(r)$ is the same as the limiting distribution of $N$ times the first term in the right hand side of \eqref{stat-2}. Note that
 $N^{1/2}w_{ij}^{1/2}E_{ij}$ converges in distribution to normal with mean $\sigma_{ij} ^{-1}\Delta_{ij}$ and variance one.
 It follows that the limiting distribution of $N$ times the first term in the right hand side of \eqref{stat-2}, and hence the limiting distribution of $T_N(r)$, is noncentral chi-square with degrees of freedom $\nu=m-\dim\left(P_\O(\LL)\right )$ and the  noncentrality parameter $\delta_r$.
  Recall that dimension of the linear space $\LL$ is equal to the sum of the dimension of its image $P_\O\left(\LL\right)$ plus the dimension of the kernel ${\rm Ker}(P_\O)$.
 It remains to note that condition \eqref{local-1} means that ${\rm Ker}(P_\O)=\{0\}$
(see Remark \ref{rem-uniq}), and hence
\begin{equation}\label{dimtan}
\dim\left(P_\O(\LL)\right)
=\dim\left(\LL\right)=r(n_1+n_2-r).
\end{equation}
This completes the proof.




\vspace{.1in}
\noindent{\bf Justification for Theorem \ref{pr-sdpuniq}}
Note that for  both problems \eqref{comp-11} and  \eqref{comp-12} the Slater condition holds, and hence there is no duality gap between these problems,  and both problems have nonempty bounded sets of optimal solutions.
 Optimality conditions (necessary and sufficient) for problem \eqref{sim-3} are
 \begin{eqnarray}
 \label{sdpopt-1}
 && C=P_{\cS^c} (\Lambda),\\
 \label{sdpopt-2}
 &&  (\Xi+X)\Lambda=0,\\
 \label{sdpopt-3}
 && \Lambda \succeq 0,\;\Xi+X \succeq 0,\; X\in \bbw_{\cS^c}.
 \end{eqnarray}
Now suppose that $\bar{X}\in \bbw_{\cS^c}$ is such that $\Xi+\bar{X}\succeq 0$ and $\rank (\Xi+\bar{X})=r<p$.
 Let $E$ be a $p\times (p-r)$ matrix of rank $p-r$ such that
 $(\Xi+\bar{X})E=0$.  By the optimality conditions \eqref{sdpopt-1}--\eqref{sdpopt-3} we have that $\bar{X}$ is an optimal solution of the SDP problem \eqref{sim-3} if and only if the following condition holds:
  there  exists $Z\in \bbs^{p-r}_+$ such that $P_{\cS^c}(EZE^\top)=C$.
 Equations $P_{\cS^c}(EZE^\top)=C$ can be viewed as a system of $\dim (\bbw_{\cS^c})$ equations with $(p-r)(p-r+1)/2$ unknowns (nonduplicated elements of matrix $Z\in \bbs^{p-r}$). When $r$ is ``small" and consequently
 $(p-r)(p-r+1)/2>\dim (\bbw_{\cS^c})$, it is likely that this system will have a solution  $Z\succeq 0$, and hence $\bar{X}$ is an optimal solution of problem \eqref{sim-3}. 
We can also view this by adjusting weight matrix $C$ to the considered matrix  $\Xi+\bar{X}$ by choosing $Z\succeq 0$ and {\em defining} $C:=P_{\cS^c}(EZE^\top)$. For such $C$ the corresponding SDP problem has $\bar{X}$ as an optimal solution. Note that although matrix $EZE^\top$ is positive semidefinite when $Z\succeq 0$, there is no guarantee that the corresponding matrix $P_{\cS^c}(EZE^\top)$ is positive semidefinite.

\end{document}